\documentclass[10pt,twocolumn,letterpaper]{article}

\usepackage{cvpr}              

\usepackage{times}
\usepackage{epsfig}
\usepackage{graphicx}
\usepackage{amsmath}
\usepackage{amssymb}
\usepackage[usenames,dvipsnames]{xcolor}
\usepackage{amsmath}
\usepackage{amssymb}
\usepackage{multirow}
\usepackage{caption}
\usepackage{cite}
\usepackage{enumitem}

\definecolor{cvprblue}{rgb}{0.21,0.49,0.74}
\usepackage[pagebackref,breaklinks,colorlinks,citecolor=cvprblue]{hyperref}

\usepackage[capitalize]{cleveref}
\crefname{section}{Sec.}{Secs.}
\Crefname{section}{Section}{Sections}
\Crefname{table}{Table}{Tables}
\crefname{table}{Tab.}{Tabs.}

\def\paperID{8189} 
\def\confName{CVPR}
\def\confYear{2024}

\begin{document}

\title{Self-Distilled Masked Auto-Encoders are Efficient 
Video Anomaly Detectors\vspace*{-0.3cm}}

\author{Nicolae-C\u{a}t\u{a}lin Ristea$^{1,2,\diamond}$, Florinel-Alin Croitoru$^{1,\diamond}$, Radu Tudor Ionescu$^{1,3,}$\thanks{corresp.~author: raducu.ionescu@gmail.com;~~$^\diamond$equal contribution}\;, Marius Popescu$^{1,3,}$\\ 
Fahad Shahbaz Khan$^{4,5}$, Mubarak Shah$^{6}$\\
$^1$University of Bucharest, Romania, $^2$NUST Politehnica Bucharest, Romania,\\
$^3$SecurifAI, Romania, $^4$MBZ University of Artificial Intelligence, UAE,\\
$^5$Link\"{o}ping University, Sweden, $^6$University of Central Florida, US\vspace*{-0.3cm}
}

\maketitle

\makeatletter
\newcommand*\mysize{%
  \@setfontsize\mysize{7.5}{9.0}%
}
\makeatother

\begin{abstract}
\vspace{-0.3cm}
We propose an efficient abnormal event detection model based on a lightweight masked auto-encoder (AE) applied at the video frame level. The novelty of the proposed model is threefold. First, we introduce an approach to weight tokens based on motion gradients, thus shifting the focus from the static background scene to the foreground objects. Second, we integrate a teacher decoder and a student decoder into our architecture, leveraging the discrepancy between the outputs given by the two decoders to improve anomaly detection. Third, we generate synthetic abnormal events to augment the training videos, and task the masked AE model to jointly reconstruct the original frames (without anomalies) and the corresponding pixel-level anomaly maps. Our design leads to an efficient and effective model, as demonstrated by the extensive experiments carried out on four benchmarks: Avenue, ShanghaiTech, UBnormal and UCSD Ped2. The empirical results show that our model achieves an excellent trade-off between speed and accuracy, obtaining competitive AUC scores, while processing 1655 FPS. Hence, our model is between 8 and 70 times faster than competing methods. We also conduct an ablation study to justify our design. Our code is freely available at: \url{https://github.com/ristea/aed-mae}.
\vspace{-0.6cm}
\end{abstract}

\setlength{\abovedisplayskip}{3.0pt}
\setlength{\belowdisplayskip}{3.0pt}

\section{Introduction}
\vspace{-0.1cm}

\begin{figure}[!t]
\begin{center}
\centerline{\includegraphics[width=1.0\linewidth]{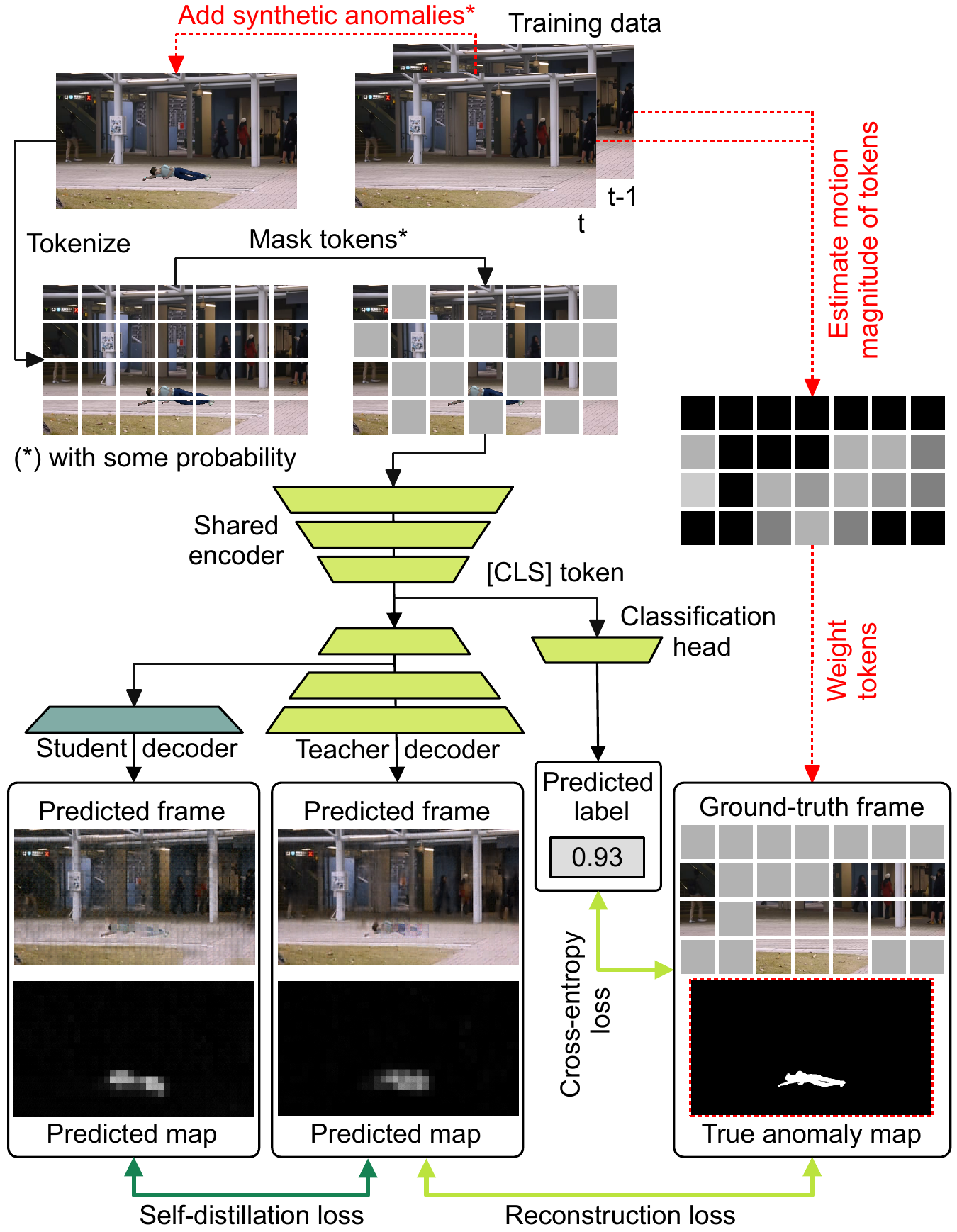}}
\vspace{-0.25cm}
\caption{Our masked auto-encoder for abnormal event detection based on self-distillation. At training time, some video frames are augmented with synthetic anomalies. The teacher decoder learns to reconstruct original frames (without anomalies) and predict anomaly maps. The student decoder learns to reproduce the teacher's output. Motion gradients are aggregated at the token level and used as weights for the reconstruction loss. Red dashed lines represent steps executed only during training.}
\label{fig_pipeline}
\vspace{-1.1cm}
\end{center}
\end{figure}

In recent years, research on abnormal event detection in video gained significant traction \cite{Acsintoae-CVPR-2022,Barbalau-CVIU-2023,Dong-Access-2020,Doshi-CVPRW-2020a,Georgescu-CVPR-2021,Georgescu-TPAMI-2021,Gong-ICCV-2019,Ionescu-CVPR-2019,Ji-IJCNN-2020,Lee-TIP-2019,Li-ECCV-2022,Liu-ICCV-2021,Lu-ECCV-2020,Nguyen-ICCV-2019,Pang-CVPR-2020,Park-CVPR-2020,Ramachandra-WACV-2020a,Ramachandra-PAMI-2020,Ristea-CVPR-2022,Smeureanu-ICIAP-2017,Sun-ACMMM-2020,Tang-PRL-2020,Vu-AAAI-2019,Wang-ACMMM-2020,Wu-TNNLS-2019,Yu-ACMMM-2020,Yu-CVPR-2022,Zaheer-CVPR-2020,Zaheer-ECCV-2020,Zaheer-CVPR-2022}, due to its utter importance in video surveillance. Despite the growing interest, video anomaly detection remains a complex task, owing its complexity to the fact that abnormal situations are context-dependent and do not occur very often. This makes it very difficult to collect a representative set of abnormal events for training state-of-the-art deep learning models in a fully supervised manner. 
To showcase the rarity and reliance on context of anomalies, we refer to the vehicle ramming attacks carried out by terrorists against pedestrians. As soon as a car is steered on the sidewalk, it becomes an abnormal event. Hence, the place where the car is driven (street versus sidewalk) determines the normal or abnormal label of the action, \ie the label depends on context. Furthermore, there are less than 200 vehicle ramming attacks registered to date\footnote{\url{https://en.wikipedia.org/wiki/Vehicle-ramming_attack}}, confirming the scarcity of such events (even less are caught on video).

Since training anomaly detectors under a fully supervised setting is not possible, most studies dealing with abnormal event detection took a distinct path, proposing variations of outlier detection methods \cite{Antic-ICCV-2011,Cheng-CVPR-2015,Cong-CVPR-2011,Dong-Access-2020,Dutta-AAAI-2015, Hasan-CVPR-2016,Ionescu-WACV-2019,Kim-CVPR-2009,Lee-TIP-2019,Li-PAMI-2014,Liu-CVPR-2018,Lu-ICCV-2013,Luo-ICCV-2017,Mahadevan-CVPR-2010,Mehran-CVPR-2009,Park-CVPR-2020,Ramachandra-WACV-2020a,Ramachandra-WACV-2020b,Ramachandra-MVA-2021,Ramachandra-PAMI-2020,Ravanbakhsh-WACV-2018,Ravanbakhsh-ICIP-2017,Ren-BMVC-2015,Sabokrou-IP-2017,Tang-PRL-2020,Wu-TNNLS-2019,Xu-CVIU-2017,Zhao-CVPR-2011,Zhang-PR-2020, Zhang-PR-2016, Zhong-CVPR-2019, Fan-CVIU-2020}. Such methods treat abnormal event detection as an outlier detection task, where a normality model trained on normal events is applied on both normal and abnormal events during inference, labeling events deviating from the learned model as abnormal. Different from the mainstream path based on outlier detection, we propose an approach to augment each training video scene with synthetic anomalies, by randomly superimposing temporal action segments from the synthetic UBnormal data set \cite{Acsintoae-CVPR-2022} on our real-world data sets. We thus introduce synthetic anomalies at training time, enabling our model to learn in an open-set supervised manner. Additionally, we force our model to reconstruct the original training frames (without anomalies) to limit its ability to reconstruct anomalies, hence generating higher errors when anomalies occur.

\begin{figure}[!t]
\begin{center}
\centerline{\includegraphics[width=1.0\linewidth]{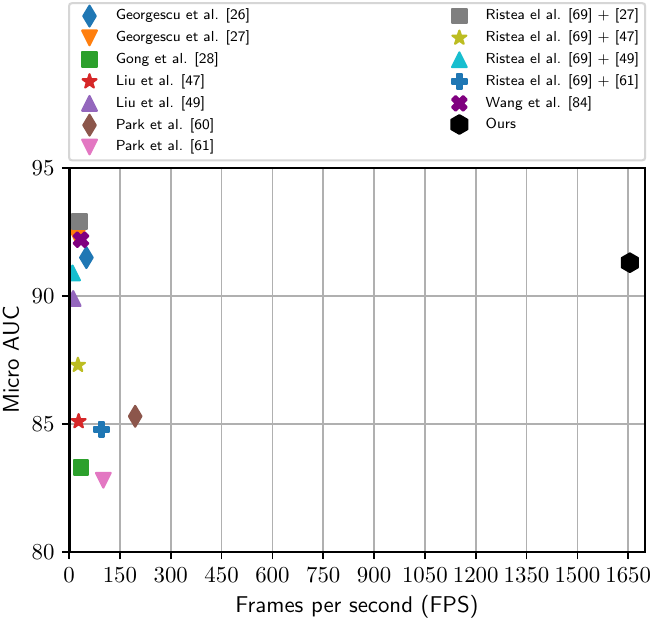}}
\vspace{-0.25cm}
\caption{Performance versus speed trade-offs for our self-distilled masked AE and several state-of-the-art methods \cite{Georgescu-CVPR-2021,Georgescu-TPAMI-2021,Ristea-CVPR-2022,Liu-CVPR-2018,Liu-ICCV-2021,Park-CVPR-2020,Gong-ICCV-2019,Wang-ECCV-2022,Park-WACV-2022} (with open-sourced code), on the Avenue data set. The running times of all methods are measured on a computer with one Nvidia GeForce GTX 3090 GPU with 24 GB of VRAM. Best viewed in color.}
\label{fig_tradeoff_av}
\vspace{-0.9cm}
\end{center}
\end{figure}

A large body of work on video anomaly detection has focused on employing auto-encoders (AEs) to address the task \cite{Astrid-ICCVW-2021,Fan-CVIU-2020,Georgescu-TPAMI-2021,Gong-ICCV-2019,Hasan-CVPR-2016,Ionescu-CVPR-2019,Liu-ICCV-2021,Tran-BMVC-2017,Wang-ICIP-2018}, relying on the poor reconstruction capabilities of these models on out-of-distribution data. Since training is carried out only on normal examples, it is expected for AEs to exhibit high reconstruction errors when anomalies occur. However, several researchers observed that AEs generalize too well \cite{Astrid-ICCVW-2021,Ionescu-CVPR-2019}, being able to reconstruct anomalies with very high precision. Thus, to better leverage the reconstruction error of AEs in anomaly detection, researchers explored a few alternatives, from the use of dummy \cite{Ionescu-CVPR-2019} or pseudo-anomalies \cite{Astrid-ICCVW-2021,Georgescu-TPAMI-2021} to the integration of memory modules \cite{Gong-ICCV-2019,Liu-ICCV-2021,Park-CVPR-2020}. With the same purpose in mind, we propose to employ masked auto-encoders \cite{He-CVPR-2022} in anomaly detection, introducing new ways to regulate their generalization capacity. Indeed, we go beyond employing the standard masked AE framework, and propose three novel changes to enhance the anomaly detection performance of our model. First, we propose to weight tokens based on the magnitude of motion gradients, raising the importance of tokens with higher motion in the reconstruction loss. This makes our model focus on reconstructing tokens with high motion, and avoid reconstructing the background scene, which is typically static for surveillance cameras. Second, we attach a classification head to discriminate between normal and pseudo-abnormal instances in the latent encoding space. Third, we integrate a teacher decoder and a student decoder into our masked AE architecture, where the student decoder learns to distill knowledge from the already optimized teacher. To reduce our processing time, we use a shared encoder for the teacher and student models, leading to a process known as self-distillation \cite{Zhang-PAMI-2022}. During the self-distillation process, the shared encoder is frozen. We leverage the discrepancy between the outputs given by the teacher and student decoders along with the reconstruction error of the teacher to boost anomaly detection performance. Our entire framework, integrating these components into a meticulous design, is shown in Figure \ref{fig_pipeline}.

State-of-the-art deep anomaly detectors \cite{Barbalau-CVIU-2023,Georgescu-CVPR-2021,Georgescu-TPAMI-2021,Wang-ECCV-2022} typically rely on a costly object detection method to increase precision, limiting the processing bandwidth to one video stream per GPU, at around 20-30 FPS. However, for real-world video surveillance, \eg monitoring an entire city with hundreds or thousands of cameras, the processing costs of object-centric video anomaly detectors are simply too high, given their power consumption and that one GPU can cost around \$2,000. To this end, we turn our attention to developing a lightweight model (6 transformer blocks, 3M parameters), capable of processing around 66 video streams at 25 FPS, significantly reducing the processing costs. Different from competing models performing anomaly detection at the object \cite{Barbalau-CVIU-2023,Doshi-CVPRW-2020a,Georgescu-CVPR-2021,Georgescu-TPAMI-2021,Ionescu-CVPR-2019,Wang-ECCV-2022,Yu-ACMMM-2020} or spatio-temporal cube \cite{Dutta-AAAI-2015,Kim-CVPR-2009,Lu-ICCV-2013,Giorno-ECCV-2016,Ionescu-ICCV-2017,Ionescu-WACV-2019,Park-WACV-2022,Liu-BMVC-2018,Luo-ICCV-2017,Mahadevan-CVPR-2010,Sabokrou-IP-2017,Saligrama-CVPR-2012,Zhang-PR-2016} levels, we present a model that takes whole video frames as input, which is significantly more efficient (see Figure \ref{fig_tradeoff_av}).

We carry out comprehensive experiments on four benchmarks: Avenue \cite{Lu-ICCV-2013}, ShanghaiTech \cite{Luo-ICCV-2017}, UBnormal \cite{Acsintoae-CVPR-2022} and UCSD Ped2 \cite{Mahadevan-CVPR-2010}. The empirical results show that our method is 8 to 70 times faster than competing methods \cite{Georgescu-CVPR-2021,Georgescu-TPAMI-2021,Ristea-CVPR-2022,Liu-CVPR-2018,Liu-ICCV-2021,Park-CVPR-2020,Gong-ICCV-2019,Wang-ECCV-2022,Park-WACV-2022}, while achieving comparable accuracy levels. Aside from the main results, we conduct an ablation study showing that our novel design choices are supported by empirical evidence.

In summary, our contribution is threefold:
\begin{itemize}
    \item \vspace{-0.00cm} We propose a lightweight masked auto-encoder for anomaly detection in video, which learns to reconstruct tokens with higher motion magnitude.    
    \item \vspace{-0.00cm} We introduce a self-distillation training pipeline, leveraging the discrepancy between teacher and student decoders to obtain a significant accuracy boost for our highly efficient model (due to the shared encoder).
    \item \vspace{-0.00cm} To further boost the performance of our model, we introduce a data augmentation approach based on superimposing synthetic anomalies on normal training videos, which enables the masked AE model to learn with open-set supervision.
\end{itemize}

\vspace{-0.15cm}
\section{Related Work}
\vspace{-0.1cm}

Anomaly detection in video is typically formulated as a one-class learning problem, where only normal data is available at training time. During test time, both normal and abnormal examples are present \cite{Pang-CSUR-2021,Ramachandra-PAMI-2020}. There are several categories of anomaly detection approaches, including dictionary learning methods \cite{Cheng-CVPR-2015,Cong-CVPR-2011,Dutta-AAAI-2015,Lu-ICCV-2013,Ren-BMVC-2015,Wu-ECCV-2022}, probabilistic models \cite{Adam-PAMI-2008,Antic-ICCV-2011,Feng-NC-2017,Hinami-ICCV-2017,Kim-CVPR-2009,Mahadevan-CVPR-2010,Mehran-CVPR-2009,Saleh-CVPR-2013,Wu-CVPR-2010}, change detection frameworks \cite{Giorno-ECCV-2016,Ionescu-ICCV-2017,Liu-BMVC-2018,Pang-CVPR-2020}, distance-based models \cite{Ionescu-CVPR-2019,Ionescu-WACV-2019,Ramachandra-WACV-2020a,Ramachandra-WACV-2020b,Ravanbakhsh-WACV-2018,Sabokrou-IP-2017,Sabokrou-CVIU-2018,Saligrama-CVPR-2012,Smeureanu-ICIAP-2017,Sun-PR-2017,Tran-BMVC-2017} and reconstruction-based approaches \cite{Fei-TMM-2020,Gong-ICCV-2019,Hasan-CVPR-2016,Li-BMVC-2020,Liu-CVPR-2018,Luo-ICCV-2017,Nguyen-ICCV-2019,Park-CVPR-2020,Ravanbakhsh-ICIP-2017,Ristea-CVPR-2022,Tang-PRL-2020}. Considering that reconstruction-based methods often reach state-of-the-art performance in anomaly detection \cite{Georgescu-TPAMI-2021,Ristea-CVPR-2022}, a large body of works used the reconstruction-based paradigm in the past few years. To this end, we adopt this paradigm in our study.

With respect to the level at which the anomaly detection is carried out, methods can be categorized into spatio-temporal cube-level methods \cite{Liu-CVPR-2018,Mehran-CVPR-2009,Ravanbakhsh-WACV-2018,Ravanbakhsh-ICIP-2017,Smeureanu-ICIAP-2017,Yu-TNNLS-2021,Ionescu-WACV-2019,Liu-BMVC-2018,Lu-ICCV-2013,Mahadevan-CVPR-2010,Sabokrou-IP-2017,Saligrama-CVPR-2012,Zhang-PR-2016}, frame-level methods \cite{Liu-CVPR-2018,Ravanbakhsh-ICIP-2017,Ravanbakhsh-WACV-2018}, and object-level methods \cite{Doshi-CVPRW-2020a,Doshi-CVPRW-2020b,Georgescu-CVPR-2021,Georgescu-TPAMI-2021,Liu-ICCV-2021,Ionescu-CVPR-2019,Wang-ECCV-2022,Yu-ACMMM-2020, Barbalau-CVIU-2023}. 




\noindent
\textbf{Frame-level and cube-level methods.} Before the deep learning era, preliminary abnormal event detection models commonly relied on taking short video sequences and dividing them into spatio-temporal cuboids \cite{Dutta-AAAI-2015,Kim-CVPR-2009,Lu-ICCV-2013,Giorno-ECCV-2016,Mahadevan-CVPR-2010,Saligrama-CVPR-2012,Zhang-PR-2016}. The cubes are then considered as independent examples, being passed as input to a machine learning model. This mainstream practice continued during the deep learning period \cite{Hasan-CVPR-2016,Ionescu-ICCV-2017,Ionescu-WACV-2019,Lee-TIP-2019,Li-ECCV-2022,Luo-ICCV-2017,Park-WACV-2022,Ramachandra-WACV-2020a,Ramachandra-WACV-2020b,Sabokrou-IP-2017,Smeureanu-ICIAP-2017,Zaheer-CVPR-2022}, when deep networks have been used to extract features \cite{Ionescu-ICCV-2017,Ionescu-WACV-2019,Liu-BMVC-2018,Luo-ICCV-2017} or learn \cite{Gong-ICCV-2019,Hasan-CVPR-2016,Lee-TIP-2019,Li-ECCV-2022,Luo-ICCV-2017,Park-WACV-2022,Ramachandra-WACV-2020a,Ramachandra-WACV-2020b,Ramachandra-MVA-2021,Sabokrou-IP-2017,Wu-ECCV-2022,Zaheer-CVPR-2022} from these spatio-temporal cubes. 

At the same time, some studies considered using entire video frames as input \cite{Liu-CVPR-2018,Ravanbakhsh-ICIP-2017,Ravanbakhsh-WACV-2018}. For example, Liu \etal~\cite{Liu-CVPR-2018} proposed an effective algorithm, which learns to reconstruct the next frame of a short video sequence. A more complex approach is proposed by Ravanbakhsh \etal~\cite{Ravanbakhsh-WACV-2018}, who employ optical-flow reconstruction to predict the anomalous regions from an input image. In a different study, Ravanbakhsh \etal~\cite{Ravanbakhsh-ICIP-2017} proposed to detect anomalies at the frame level via generative adversarial networks.

Frame-level and cube-level methods have a common characteristic, namely their relatively high processing speed due to the reasonably fast preprocessing steps, as opposed to object-centric methods. Still, frame-level methods hold a stronger advantage in terms of time, since cube-level methods need to process each cube as an independent example. Indeed, it is more efficient to process a mini-batch of frames rather than several mini-batches of spatio-temporal cubes. However, cube-level methods often outperform frame-level methods. To this end, we propose a masked AE that takes whole frames as input, yet learns interactions between video patches, thus integrating the best of both worlds.


To boost the performance of frame-level or cube-level methods, researchers explored the inclusion of various components, such as memory modules \cite{Gong-ICCV-2019,Park-CVPR-2020} or masked convolutional blocks \cite{Ristea-CVPR-2022}. Although integrating additional modules into the framework leads to accuracy gains, the procedure often comes with efficiency drawbacks. In contrast, our goal is to achieve a superior trade-off between performance and speed, with a higher focus on efficiency. As such, we design a lightweight masked AE based on convolutional vision transformer (CvT) blocks \cite{Wu-ICCV-2021}, and propose several upgrades resulting in a minimal time overhead. For example, we employ knowledge distillation to leverage the discrepancy between the teacher and the student models. However, to keep the processing time to the bare minimum, we resort to self-distillation \cite{Zhang-PAMI-2022} and use a shared encoder for the teacher and student networks.

\noindent
\textbf{Object-level methods.} To reduce the number of false positive detections often observed for other methods, some recent studies \cite{Barbalau-CVIU-2023, Ionescu-CVPR-2019, Georgescu-TPAMI-2021, Georgescu-CVPR-2021,Liu-ICCV-2021,Yu-ACMMM-2020,Wang-ECCV-2022} proposed to look for anomalous objects rather than anomalous frames or cubes. Object-centric methods use the prior information from an object detector, enabling the anomaly detector to focus only on objects. This kind of framework boosts the accuracy by significant margins, currently reaching state-of-the-art performance \cite{Barbalau-CVIU-2023,Wang-ECCV-2022}.
However, a considerable drawback is that 
the inference speed of the whole framework is directly conditioned by the object detector's speed, 
which is often much lower than that of the anomaly detection network \cite{Ionescu-CVPR-2019,Georgescu-CVPR-2021}. Hence, the processing time is significantly limited. In contrast, we perform anomaly detection at the frame level, obtaining an inference speed that is between 32 to 70 times faster than object-centric models \cite{Barbalau-CVIU-2023, Ionescu-CVPR-2019, Georgescu-TPAMI-2021, Georgescu-CVPR-2021,Liu-ICCV-2021,Yu-ACMMM-2020,Wang-ECCV-2022}.

\noindent
\textbf{Masked auto-encoders in anomaly detection.}
He \etal~\cite{He-CVPR-2022} proposed masked auto-encoders as a pretraining method to obtain strong backbones for downstream tasks. Since then, the method has been adopted in various fields, \eg~video processing \cite{Feichtenhofer-2022-NIPS} or multimodal learning \cite{Bachmann-ECCV-2022}, with remarkable results. We elaborate the connection to seemingly related masked AEs in the supplementary \cite{Chen-ECCV-2022,Yang-ARXIV-2022}.
The masking framework has also been used for anomaly detection in medical \cite{Huang-TM-2022} and industrial \cite{Jiang-TII-2022} images. To the best of our knowledge, we are the first to propose a masked transformer-based auto-encoder for video anomaly detection. Moreover, we go beyond applying standard masked AEs, proposing several modifications leading to superior performance levels: emphasizing tokens with higher motion, augmenting training videos with synthetic anomalies, and employing self-distillation.


\noindent
\textbf{Knowledge distillation in anomaly detection.}
Knowledge distillation \cite{Ba-NIPS-2014,Hinton-DLRL-2015} 
was originally designed to compress one or multiple large models (teachers) into a lighter neural network (student). Recently adopted in anomaly detection \cite{Bergmann-CVPR-2020,Cheng-PRCV-2021,Deng-CVPR-2022,Georgescu-CVPR-2021,Salehi-CVPR-2021,Wang-MLSP-2021}, knowledge distillation was deemed useful due to the possibility of leveraging the representation discrepancy between the teacher and the student networks, which is larger in the case of anomalies. For example, Bergmann \etal~\cite{Bergmann-CVPR-2020} trained an ensemble of student networks on normal data to reproduce the output of a deep feature extractor, which is pretrained on ImageNet \cite{Russakovsky-IJCV-2015}. The authors use the difference between the teacher label and the mean over student labels to detect abnormal pixels. Salehi \etal~\cite{Salehi-CVPR-2021} employed a more thorough distillation process, called hint learning, in which the multi-level features of a teacher pretrained on ImageNet are distilled into a clone. 

Most studies based on knowledge distillation applied the framework to image anomaly detection \cite{Bergmann-CVPR-2020,Cheng-PRCV-2021,Deng-CVPR-2022,Salehi-CVPR-2021}. With few exceptions \cite{Georgescu-CVPR-2021,Wang-MLSP-2021}, knowledge distillation in video anomaly detection remains  largely unexplored. Wang et al.~\cite{Wang-MLSP-2021} employed the teacher-student training paradigm to learn from unlabeled video samples in a self-supervised manner. Georgescu \etal~\cite{Georgescu-CVPR-2021} integrated knowledge distillation as a proxy task into a multi-task learning framework for video anomaly detection. 

Distinct from the aforementioned studies, to our knowledge, we are the first to introduce a variant of self-distillation in anomaly detection. Self-distillation \cite{Zhang-PAMI-2022} attaches multiple classification heads at various depths to boost the classification performance of a neural classifier. In contrast, we integrate self-distillation into a masked AE, employing two decoders of different depths. Due to the shared encoder, we are able to leverage the reconstruction discrepancy between the teacher and the student with a minimal computational overhead.

\vspace{-0.15cm}
\section{Method}
\vspace{-0.1cm}

\noindent
\textbf{Overview.}
We introduce a lightweight teacher-student transformer-based masked AE, which employs a two-stage training pipeline. In the first stage, we optimize a teacher masked AE via a reconstruction loss that employs a novel weighting mechanism based on motion gradients. In the second stage, we optimize the last (and only) decoder block of a student masked AE, which shares most of the backbone (kept frozen) with its teacher, to preserve efficiency.
Next, we describe how to create training videos with synthetic anomalies and train the masked AEs to jointly predict the anomaly maps and overlook (not reconstruct) the anomalies from training frames. Lastly, we introduce a classification head to distinguish between frames with and without synthetic anomalies, which further boosts the performance of our method, with a marginal computational overhead.

\noindent
\textbf{Architecture.}
Our masked AE pursues the architectural principles proposed in \cite{He-CVPR-2022}. Hence, the entire architecture is formed of visual transformer blocks. In contrast to He \etal~\cite{He-CVPR-2022}, we replace the ViT \cite{Dosovitskiy-ICLR-2020} blocks with CvT blocks \cite{Wu-ICCV-2021}, aiming for higher efficiency. Our processing starts by dividing the input images into non-overlapping tokens and removing a certain number of tokens. The encoder embeds the remaining tokens via convolutional projection layers, and the result is processed by transformer blocks. The decoder operates on a complete set of tokens, those removed being replaced with mask tokens. Its architecture is symmetric to that of the encoder. For efficiency reasons, we only use three blocks for the encoder and three blocks for the decoder. Each block is equipped with four attention heads. To achieve further speed gains, we replace all dense layers inside the CvT blocks with pointwise convolutions. We consider the architecture described so far as a teacher network. A student decoder branches out from the teacher after the first transformer block of the main decoder, adding only one extra transformer block (as shown in Figure \ref{fig_pipeline}).

\noindent
\textbf{Motion gradient weighting.}
Masked AEs \cite{He-CVPR-2022} have been originally applied on natural images. In this context, reconstructing randomly masked tokens is a viable solution, since images have high foreground and background variations. However, abnormal event detection data sets \cite{Adam-PAMI-2008,Lu-ICCV-2013,Luo-ICCV-2017,Mahadevan-CVPR-2010} contain videos from fixed cameras with static backgrounds \cite{Ramachandra-PAMI-2020}. Learning to reconstruct the static background via masked AEs is both trivial and useless. Hence, naively training masked AEs to reconstruct randomly masked tokens in video anomaly detection is suboptimal. To this end, we propose to take into account the magnitude of the motion gradients when computing the reconstruction loss.

Let $\pmb{x}_t \in \mathbb{R}^{h\times w\times c}$ be the video frame at index $t$.
Let $n$ be the number of non-overlapping visual tokens (patches) of size $d\times d \times c$ from each frame $\pmb{x}_t$, where $c$ is the number of input channels, and $d$ is a hyperparameter that directly determines $n$. Let $\left\{\pmb{p}_i^{(t)}\right\}_{i=1}^{n} \in \mathbb{R}^{d\times d\times c}$ denote the set of tokens in frame $\pmb{x}_t$, and $\left\{\hat{\pmb{p}}_i^{(t)}\right\}_{i=1}^{n} \in \mathbb{R}^{d\times d\times c}$ the corresponding set of reconstructed tokens. 

Following Ionescu \etal \cite{Ionescu-CVPR-2019}, we estimate the motion gradient map $\pmb{g}_t$ of frame $\pmb{x}_t$ by computing the absolute difference between consecutive frames, which are previously filtered with a $3\times 3$ median filter. 
Next, we divide the gradient magnitude map $\pmb{g}_t$ into non-overlapping patches, obtaining the set of gradient patches $\left\{\pmb{r}_i^{(t)}\right\}_{i=1}^{n} \in \mathbb{R}^{d\times d\times c}$. Inside each gradient patch, we compute the maximum gradient magnitude per channel. Then, we compute the channel-wise mean over the maximum gradient magnitudes, as follows:
\begin{equation}
\label{max_pooling}
m_i^{(t)}=\frac{1}{c}\sum_{l=1}^{c}{\max_{{j,k}} \left\{\pmb{r}_{ijkl}^{(t)}\right\}}, \forall j, k \in \{1,...,d\}. 
\end{equation}
Finally, we compute the token-wise weights for the reconstruction loss
as follows:
\begin{equation}
 \label{eq_gradient_weights}
    \begin{split}
        w_i^{(t)} = \frac{m_i^{(t)}}{\sum_{j=1}^{n} m_{j}^{(t)}}, \forall i \in \{1,...,n\}. 
    \end{split}
\end{equation}
Introducing the resulting weights $w_i^{(t)}$ into the conventional token-level reconstruction loss leads to an objective that pushes the masked AE to focus on reconstructing the patches with high motion magnitude. Formally, our weighted mean squared error loss is given by:
\begin{equation}
 \label{eq_loss_mae}
    \begin{split}
        \mathcal{L}_{\mbox{\scriptsize{wMSE}}} (\pmb{x}_t, \theta_T) = \frac{1}{n} \sum_{i=1}^{n} w_i^{(t)} \cdot \lVert \pmb{p}_i^{(t)}-\hat{\pmb{p}}_i^{(t)}\rVert^2_2,
    \end{split}
\end{equation}
where $\theta_T$ are the weights our teacher masked AE.
Although our reconstruction loss focuses on tokens with high motion, the masked tokens are still chosen randomly.

\noindent
\textbf{Self-distillation.} Knowledge distillation has already shown its utility in anomaly detection \cite{Bergmann-CVPR-2020,Cheng-PRCV-2021,Deng-CVPR-2022,Georgescu-CVPR-2021,Salehi-CVPR-2021}. Intuitively, since the teacher and student models are both trained on normal data, their reconstructions should be very similar for normal test samples. However, their behavior is not guaranteed to be similar on abnormal examples. Therefore, the magnitude of the teacher-student output gap (discrepancy) can serve as a means to quantify the anomaly level of a given sample. Unfortunately, this approach implies using both teacher and student models during inference, virtually splitting our processing speed in half. To slash the additional burden of using another model during inference, we propose to employ a novel variant of self-distillation with a shared encoder and two decoders, a teacher and a student. 
More precisely, the student branches out from the original architecture after the first transformer block of the teacher decoder, essentially adding only one transformer block. 

Our training process is carried out in two stages. In the first phase, the teacher is trained with the loss defined in Eq.~\eqref{eq_loss_mae}. In the second phase, we freeze the weights of the shared backbone and train only the student decoder via self-distillation. The self-distillation loss is similar to the one defined in Eq.~\eqref{eq_loss_mae}. The main difference is that instead of reconstructing the patches from the real image, the student learns to reconstruct the ones produced by the teacher. Let $\left\{\tilde{\pmb{p}}_i^{(t)}\right\}_{i=1}^{n} \in \mathbb{R}^{d\times d\times c}$ denote the patches reconstructed by the student. Then, the self-distillation loss can be expressed as follows:
\begin{equation}
 \label{eq_distillation}
    \begin{split}
        \mathcal{L}_{\mbox{\scriptsize{SD}}} (\hat{\pmb{x}}_t, \theta_S) = \frac{1}{n} \sum_{i=1}^{n} w_i^{(t)} \cdot \lVert \hat{\pmb{p}}_i^{(t)}-\tilde{\pmb{p}}_i^{(t)}\rVert^2_2,
    \end{split}
\end{equation}
where $\hat{\pmb{x}}_t$ is the frame reconstructed by the teacher, and $\theta_S$ are the weights of the student decoder. Notice that we keep the motion gradient weights $w_i^{(t)}$ during self-distillation.

\begin{figure}[!t]
\begin{center}
\centerline{\includegraphics[width=0.8\linewidth]{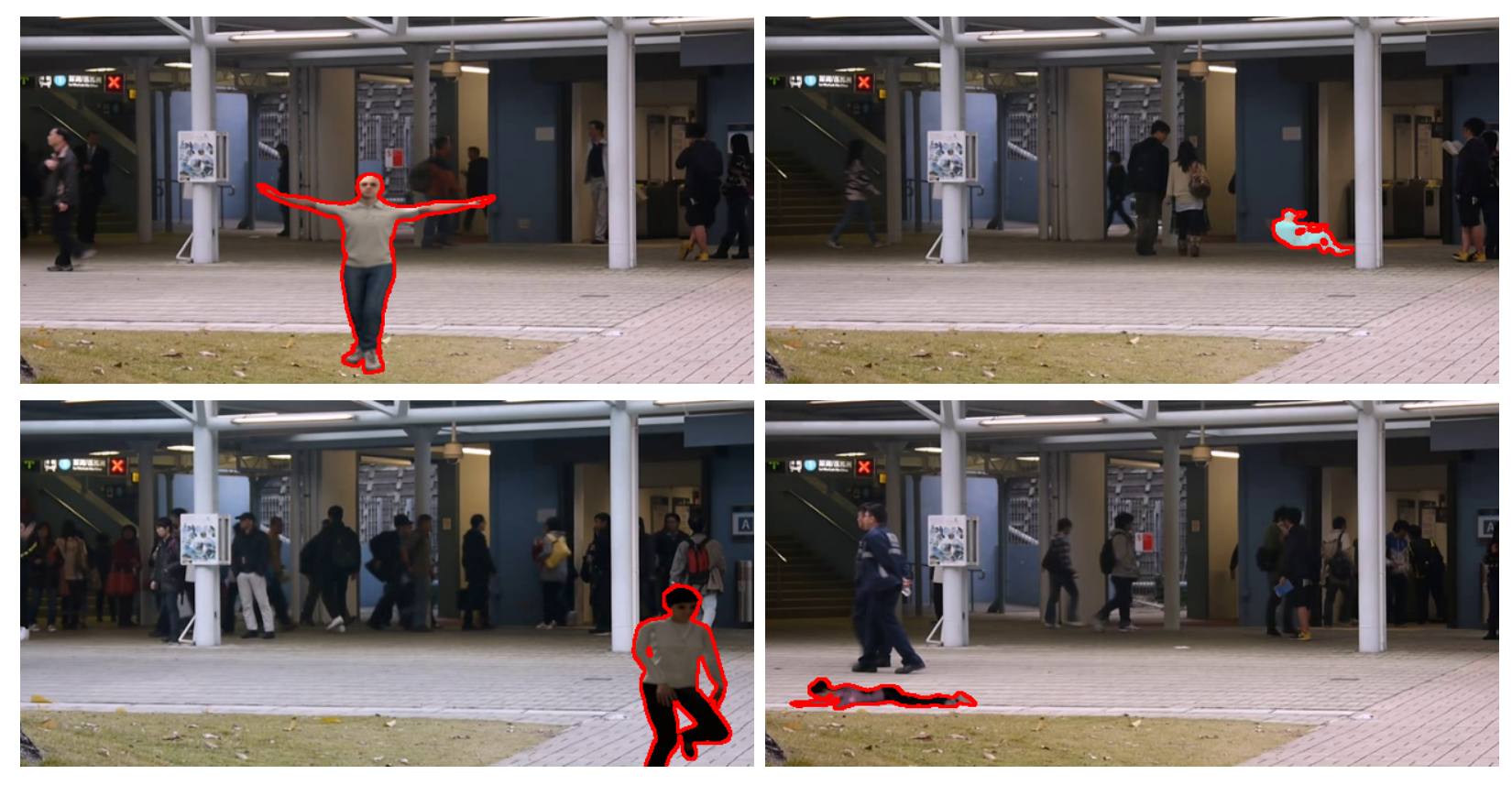}}
\vspace{-0.3cm}
\caption{Four synthetic anomalies (with red contours) taken from the UBnormal data set \cite{Acsintoae-CVPR-2022} and overlaid on training frames from Avenue \cite{Lu-ICCV-2013}. Best viewed in color.}
\label{fig_train_anomalies}
\vspace{-1.0cm}
\end{center}
\end{figure}

\noindent
\textbf{Synthetic anomalies.} As observed in other studies \cite{Astrid-ICCVW-2021,Ionescu-CVPR-2019}, AEs tend to generalize too well to out-of-distribution data. This behavior is not desired in anomaly detection, since methods based on AEs rely on having high reconstruction errors for abnormal examples and low reconstruction errors for normal ones. To this end, we propose to augment the training videos with abnormal events. Since collecting abnormal training examples from the real-world is not possible, we resort to adding synthetic (virtual) anomalies. We leverage the recently introduced UBnormal data set \cite{Acsintoae-CVPR-2022} and its accurate pixel-level annotations to crop out abnormal events and blend them in our training videos, while ensuring the temporal consistency of the added events. The resulting examples, some depicted in Figure \ref{fig_train_anomalies}, are used to augment the training set with extra data. 

The synthetic examples help our model in three ways. First, in the reconstruction loss, we consider the original training frames (without superimposed anomalies) as the ground-truth, essentially forcing our model to overlook the anomalies. Formally, in Eq.~\eqref{eq_loss_mae}, we use the patches $\left\{\pmb{p}_i^{(t)}\right\}_{i=1}^{n}$ from the normal version of frame $\pmb{x}_t$.
Second, we add the anomaly map as an additional channel to our target image. In the anomaly map, we set normal pixels to $0$ and abnormal pixels to $1$. This change implies that, in Eq.~\eqref{eq_loss_mae} and Eq.~\eqref{eq_distillation}, all patches will have an additional channel. 
Third, we use the ground-truth anomaly map to enhance the weights defined in Eq.~\eqref{eq_gradient_weights}. The added synthetic anomalies do not necessarily yield motion gradients with high magnitude. Hence, it is possible to have low weights in Eq.~\eqref{eq_loss_mae} and Eq.~\eqref{eq_distillation} for patches that correspond to anomaly regions. This is not desirable if we want the model to detect anomalies. To this end, we propose to add the anomaly maps and the gradients together, before computing the weights as in Eq.~\eqref{eq_gradient_weights}. Formally, in Eq.~\eqref{max_pooling}, we replace 
$\left\{\pmb{r}_i^{(t)}\right\}_{i=1}^n$ 
with $\left\{\pmb{r}_i^{(t)}+\pmb{a}_i^{(t)}\right\}_{i=1}^{n}$, 
where $\left\{\pmb{a}_i^{(t)}\right\}_{i=1}^n$ is the set of the patches extracted from the anomaly map.

\noindent
\textbf{Classification head.} We further harness the synthetic anomalies to train a classification head applied on the final {[CLS]} token of the shared encoder. The head is trained to discriminate between frames with and without synthetic anomalies. This head is trained using binary cross-entropy:
\begin{equation}
 \label{eq_cross_entropy}
    \begin{split}
        \mathcal{L}_{\mbox{\scriptsize{CE}}} (\hat{\pmb{x}}_t, \theta_E) = - y_t\!\cdot\! \log(\hat{y}_t) - (1\!-\!y_t)\!\cdot\!\log(1-\hat{y}_t),
    \end{split}
\end{equation}
where $y_t \in \{0,1\}$ is $1$ if the frame contains an anomaly and $0$ otherwise, $\hat{y}_t$ is the prediction, and $\theta_E$ represents the set of weights of the shared encoder.

\noindent
\textbf{Inference.} During inference, we pass each frame $\pmb{x}_{t}$ through both teacher and student models to obtain the reconstructed frames $\hat{\pmb{x}}_{t}$ and $\tilde{\pmb{x}}_{t}$, respectively. Then, we compute output pixel-level anomaly map as:
\begin{equation}\label{eq_infer}
\pmb{o}_{t} = \alpha\!\cdot\!\lVert \pmb{x}_{t}-\hat{\pmb{x}}_{t}\rVert^2_2 + \beta\!\cdot\!\lVert \hat{\pmb{x}}_{t}-\tilde{\pmb{x}}_{t}\rVert^2_2 + \gamma\!\cdot\!\hat{y}_t,
\end{equation}
where $\alpha$, $\beta$ and $\gamma$ are hyperparameters that control the importance of the individual anomaly score components. Following \cite{Giorno-ECCV-2016,Ionescu-ICCV-2017}, we apply spatio-temporal 3D filtering to smooth the anomaly volumes. To obtain the frame-level anomaly scores, we keep the maximum value from each output map $\pmb{o}_{t}$ and subsequently apply another temporal Gaussian filter to smooth the values.

\vspace{-0.15cm}
\section{Experiments}
\vspace{-0.1cm}
\subsection{Experimental Setup}
\vspace{-0.1cm}

\noindent
\textbf{Data sets.}
We verify the performance of our method on four data sets for video anomaly detection: Avenue \cite{Lu-ICCV-2013}, ShanghaiTech \cite{Luo-ICCV-2017}, UBnormal \cite{Acsintoae-CVPR-2022} and UCSD Ped2 \cite{Mahadevan-CVPR-2010}. ShanghaiTech is the largest data set, with 270K frames for training and about 50K for testing. UBnormal is the second largest, with about 116K training frames and 93K testing frames. Avenue is a popular benchmark containing 15K frames for training and another 15K for testing. UCSD Ped2 holds a total of 4.5K frames, out of which 2.5K are used for training. 
UBnormal is a benchmark that uses an open set evaluation, where training and test anomalies belong to disjoint category sets. For the other three data sets, the training videos contain only normal events, and the test ones include both normal and abnormal scenarios. To augment the normal training videos, we sample abnormal events from the UBnormal data set \cite{Acsintoae-CVPR-2022}. UBnormal is a synthetic (virtual) data set containing anomalies simulated by video game characters, which alleviates the burden of collecting anomalies from the real world. The probability of augmenting a frame from Avenue, ShanghaiTech or UCSD Ped2 is $0.25$.

\begin{table}[t]
\centering 
\setlength\tabcolsep{2.6pt}
\mysize{
\begin{tabular}{| c | c | c | c | c | c | c | c | c | c |  c |} 
\hline
 \multirow{3}{*}{\rotatebox[origin=c]{90}{\hspace{-0.5cm}Type}} &
 \multirow{3}{*}{\vspace{-0.5cm}Method} & \multicolumn{2}{c|}{Avenue} & \multicolumn{2}{c|}{Shanghai} & \multicolumn{2}{c|}{UBnormal}  & \multicolumn{2}{c|}{Ped2}  &
 \multirow{3}{*}{\vspace{-0.5cm}FPS} \\
 \cline{3-10}
 & & \multicolumn{8}{|c|}{AUC}  &\\
 \cline{3-10}
 & & \rotatebox[origin=c]{90}{$\;$Micro$\;$} & \rotatebox[origin=c]{90}{Macro} & \rotatebox[origin=c]{90}{Micro} & \rotatebox[origin=c]{90}{Macro} & \rotatebox[origin=c]{90}{Micro} & \rotatebox[origin=c]{90}{Macro} & \rotatebox[origin=c]{90}{Micro} & \rotatebox[origin=c]{90}{Macro} & \\
 \hline 
 \hline
 
 {\multirow{16}{*}{\rotatebox[origin=c]{90}{Object-centric}}} 
&
\cite{Barbalau-CVIU-2023} & 91.6 & 92.5 & {\color{RoyalBlue}83.8} & {\color{ForestGreen}90.5} & {\color{ForestGreen}62.1} & {\color{red}86.5} & - & - & 20\\
 &
 \cite{Doshi-CVPRW-2020a}  & 86.4 & - & 71.6 & - & - & - & {97.8} & - & -\\
& 
\cite{Georgescu-CVPR-2021} &  91.5 & {\color{RoyalBlue}92.8} & 82.4 & {\color{RoyalBlue}90.2} & 55.4 & {\color{RoyalBlue}84.5} & 97.5 & {\color{red}99.8} & 51 \\
& 
\cite{Georgescu-TPAMI-2021} &  {\color{RoyalBlue}92.3} & 90.4 &  82.7 & 89.3 & {\color{RoyalBlue}61.3} & {\color{ForestGreen}85.6} & {\color{RoyalBlue}98.7} & {\color{ForestGreen}99.7} & 24 \\
& \cite{Hirschorn-ICCV-2023} & - & - & {\color{red}85.9} & - & {\color{red}71.8} & - & - & - & - \\

 &
 \cite{Ionescu-CVPR-2019} & 87.4 & 90.4 & 78.7 & 84.9 & - & - & 94.3 & {\color{RoyalBlue}97.8}  & -\\
 & \cite{Kanu-A-2022} & 80.2 & - & 73.7 & - & 50.7 & - & - & - & - \\ 
&
\cite{Liu-ICCV-2021} & 89.9 & {\color{ForestGreen}93.5} & 74.2 & 83.2 & - & - & {\color{red}99.3} & -& 12 \\ 

& \cite{Liu-CVPR-2023} & 91.8 & 92.3 & {\color{RoyalBlue}83.8} & 87.8 & - & - & - & - & - \\

&
\cite{Madan-ARXIV-2022} + \cite{Barbalau-CVIU-2023} & 91.6 & {92.4} &  83.6 & {\color{red}90.6} & - & - & - & - & 20\\
&
\cite{Madan-ARXIV-2022} + \cite{Georgescu-TPAMI-2021} & {\color{red}93.2} & 91.8 & 83.3 & 89.3 & - & - & - & - & 31 \\
&
\cite{Madan-ARXIV-2022} + \cite{Liu-ICCV-2021} &  89.5 & {\color{red}93.6} &  75.2 & 83.8 & - & - & - & - & 10\\ 
&
\cite{Ristea-CVPR-2022} + \cite{Georgescu-TPAMI-2021} & {\color{ForestGreen}92.9} & 91.9 & 83.6 & 89.5 & - & - & - &- & 31 \\
&
\cite{Ristea-CVPR-2022} + \cite{Liu-ICCV-2021} &  {90.9} & {92.2} & 75.5 & {83.7} & - & - & - & - & 10\\ 
 &
 \cite{Wang-ECCV-2022} & 92.2 & - & {\color{ForestGreen}84.3} & - & - & - & {\color{ForestGreen}99.0} & - & 35\\
 &
 \cite{Yu-ACMMM-2020} & 89.6 & - & 74.8 & - & - & - & 97.3 & - & -\\

\hline
{\multirow{32}{*}{\rotatebox[origin=c]{90}{Frame or cube level}}}

&
\cite{Astrid-BMVC-2021} & 87.1 & - & 75.9 & - & - & - & 96.5 & - &-\\
&
\cite{Astrid-ICCVW-2021} & 84.7 & - & 73.7 & - & - & - & {\color{red}98.4}&- &-\\
& \cite{Bertasius-ICML-2021} & - & - & - & - & {\color{red}68.5} & {\color{ForestGreen}80.3} & - & - & 37 \\
&
\cite{Gong-ICCV-2019} & 83.3 & -  & 71.2 & - & - & - & 94.1 & - & 35\\
&
\cite{Ionescu-WACV-2019} & 88.9 & - & - & - & - & - & - & - &- \\
&
\cite{Lee-TIP-2019} & 90.0 & - &- & - & - & - &96.6 & - &-\\
&
\cite{Liu-CVPR-2018} & 85.1 & 81.7 & 72.8 & 80.6 & - & - & 95.4& - & 28\\
&
\cite{Liu-BMVC-2018} & 84.4 & - & - & - & - & - & 87.5 & - &-\\
&
\cite{Madan-ARXIV-2022} + \cite{Liu-CVPR-2018} & 89.1 & 84.8 & 74.6 & {\color{ForestGreen}83.3} & - & - & - & - & 26\\
&
\cite{Madan-ARXIV-2022} + \cite{Park-CVPR-2020} & 86.4 & 86.3 & 70.6 & 80.3 & - & - & - & - & 94\\ 
&
\cite{Nguyen-ICCV-2019} & 86.9 & - &  - & - & - & - & 96.2 & - &-\\
&
\cite{Park-WACV-2022} & 85.3 & - & 72.2 & - & - & - & 96.3 & - & 195\\
&
\cite{Park-CVPR-2020} & 82.8 & {\color{RoyalBlue}86.8} & 68.3 & 79.7 & - & - & {\color{RoyalBlue}97.0} & - & 101\\
&
\cite{Ramachandra-WACV-2020a} & 72.0 & - &  - & - & - & - & 88.3 & - &-\\
&
\cite{Ramachandra-WACV-2020b} & 87.2 & - & - & - & - & - & 93.0 & - &-\\
&
\cite{Ravanbakhsh-ICIP-2017} & - & - & - & - & - & - & 93.5 & - &-\\
&
\cite{Ravanbakhsh-WACV-2018} & - & - & - & - & - & - & 88.4 & - &-\\
&
\cite{Ristea-CVPR-2022} + \cite{Liu-CVPR-2018} & 87.3 & 84.5 & 74.5 & {\color{RoyalBlue}82.9} & - & - & - & - & 26\\
&
\cite{Ristea-CVPR-2022} + \cite{Park-CVPR-2020} & 84.8 & {\color{ForestGreen}88.6} & 69.8 & 80.2 & - & - & - & - & 95 \\
&
\cite{Smeureanu-ICIAP-2017} & 84.6 & - & - & - & - & - & - & - & -\\
&
\cite{Sultani-CVPR-2018} & - & - & - & 76.5 & 50.3 & {\color{RoyalBlue}76.8} & - & - & 56 \\
&
\cite{Sun-ACMMM-2020}  & 89.6 & - & 74.7 & - & - & - & - & - & -\\
&
\cite{Tang-PRL-2020} & 85.1 & - & 73.0 & - & - & - & 96.3 & - & -\\
& \cite{Tur-ICIP-2023}  & - & - & 76.1 & - & - & - & - & - & - \\
&
\cite{Wang-ACMMM-2020} & 87.0 & - & {\color{ForestGreen}79.3} & - & - & - & - & - & -\\
&
\cite{Wu-ECCV-2022} &  - & - & {\color{red}80.4} & - & - & - & - & - & -\\
&
\cite{Wu-TNNLS-2019} & 86.6 & - & - & - & - & - & 96.9 & -  & -\\
& \cite{Yan-ICCV-2023} & {\color{RoyalBlue}90.1} & - & 78.6 & - & {\color{ForestGreen}62.7} & - & - & - & - \\
&
\cite{Yu-TNNLS-2021} & {\color{ForestGreen}90.2} & - & - & - & - & - & {\color{ForestGreen}97.3} & - & -\\
& \cite{Zaheer-CVPR-2022} & - & - & 78.9 & - & - & - & - & - & - \\

&
\cite{Zhang-PR-2016} & - & - & - & - & - & - & 91.0 & - & - \\

\cline{2-11}

&Ours & {\color{red}91.3} & {\color{red}90.9} & {\color{RoyalBlue}79.1} & {\color{red}84.7} & {\color{RoyalBlue}58.5} & {\color{red}81.4} & 95.4 & {\color{red}98.4} & 1655\\
\hline
\end{tabular}
}
\vspace{-0.2cm}
\caption{Micro and macro AUC scores (in \%) of several state-of-the-art frame-level, cube-level and object-level methods versus our self-distilled masked AE on Avenue, ShanghaiTech, UBnormal and UCSD Ped2. The top three scores for each category of methods are shown in {\color{red}red}, {\color{ForestGreen}green}, and {\color{RoyalBlue}blue}. All reported running times (including those of the baselines) are measured on a machine with an Nvidia GeForce GTX 3090 GPU with 24 GB of VRAM.}
\vspace{-0.6cm}
\label{tab_results} 
\end{table}

\noindent
\textbf{Evaluation.}
We evaluate all models following recent related works \cite{Acsintoae-CVPR-2022,Georgescu-TPAMI-2021,Ristea-CVPR-2022}, considering both micro and macro AUC metrics. The area under the ROC curve (AUC) expresses the overlap between the ground-truth frame-level annotations and the anomaly scores predicted by a model, at multiple thresholds. At a given threshold, a frame is labeled as abnormal if the predicted anomaly score is above the threshold. For the micro AUC, the test frames from all videos are concatenated before computing the AUC over all frames. For the macro AUC, the AUC of each test video is first computed, and the resulting AUC scores are averaged to obtain a single value.

\noindent
\textbf{Hyperparameters.}
The encoder module is formed of three CvT blocks, each with a projection size of 256 and four attention heads. The teacher decoder contains three CvT blocks, while the student decoder contains only one block. All decoder blocks have four attention heads and a projection dimension of 128. Since the data sets have different input resolutions and objects vary in size, we adapt the patch size to each data set. Thus, we set the patch size to $16\times16$ on Avenue, $8 \times 8$ on ShanghaiTech and UBnormal, and $4 \times 4$ on UCSD Ped2. Regardless of the data set, the teacher network is trained for 100 epochs, while the student is trained for 40 epochs. We optimize the networks with Adam \cite{Kingma-ICLR-2014}, using a learning rate of $10^{-4}$ and mini-batches of 100 samples. The hyperparameters in Eq.~\eqref{eq_infer} are set to $\alpha=0.4$, $\beta=0.3$ and $\gamma=0.3$, for all data sets.

\vspace{-0.1cm}
\subsection{Results}
\vspace{-0.1cm}

We present our results in Table \ref{tab_results} and discuss them below.

\begin{figure}[!t]
\begin{center}
\centerline{\includegraphics[width=0.9\linewidth]{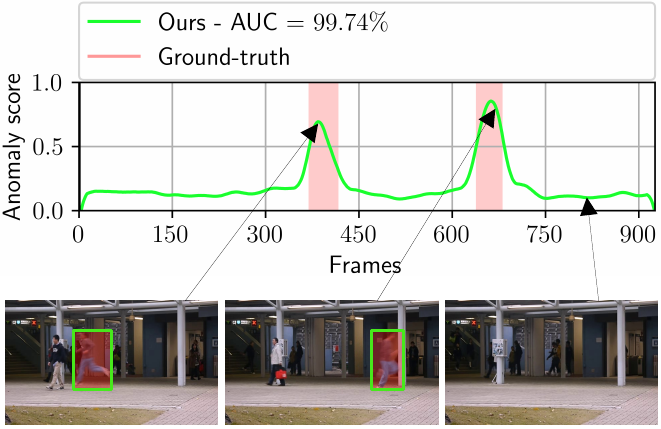}}
\vspace{-0.25cm}
\caption{Predictions for test video $04$ from Avenue. The abnormal bounding boxes are given by the convex hull of the patches labeled as abnormal. Best viewed in color.}
\label{fig_avenue_example}
\vspace{-0.9cm}
\end{center}
\end{figure}

\noindent
\textbf{Results on Avenue.}
Our method obtains a micro AUC score of $91.3\%$ on Avenue, being only $1.9\%$ below the state-of-the-art object-centric method. 
Remarkably, in the category of frame-level methods, we reach the best micro and macro AUC scores. Taking into account that our method is much faster than all other methods, we consider that its performance is remarkable. In Figure \ref{fig_avenue_example}, we illustrate the anomaly scores for test video $04$ from Avenue. Here, our model is close to perfect, highlighting its ability to capture anomalies, such as people running.

\noindent
\textbf{Results on ShanghaiTech.} On ShanghaiTech, our method reaches the top macro AUC score and the third-best micro AUC score, when compared with other frame-level frameworks. Object-centric methods generally surpass frame-level methods, but the former methods have much lower processing speeds (see Figure \ref{fig_tradeoff_av}).

\noindent
\textbf{Results on UBnormal.} In terms of the micro AUC, the best frame-level method on UBnormal is TimeSformer \cite{Bertasius-ICML-2021}, which benefits from large-scale pretraining. Notably, our method obtains a higher macro AUC than TimeSformer. The micro AUC of our method is fairly close to the micro AUC levels of the better object-centric approaches. In terms of speed, our method is significantly faster than all the other methods reporting results on UBnormal.

\noindent
\textbf{Results on UCSD Ped2.}
Our framework obtains a micro AUC of $95.4\%$, being $3.9\%$ below the state-of-the-art performance of Liu \etal~\cite{Liu-ICCV-2021}. While being slightly below in terms of micro AUC, our macro AUC score is remarkably on par with the object-centric method of Ionescu \etal~\cite{Ionescu-CVPR-2019}, and only $1.4\%$ below that of Georgescu \etal~\cite{Georgescu-CVPR-2021}. Considering that our FPS is more than $30$ times higher compared with these methods \cite{Georgescu-CVPR-2021,Ionescu-CVPR-2019} on the same GPU, our framework provides a clearly superior accuracy-speed trade-off.

\begin{table}[t]
\centering 
\setlength\tabcolsep{1.45pt}
\mysize{
\begin{tabular}{| c | c | c | c | c | c | c | c | c |} 
\hline
   \multirow{2}{*}{Motion} & \multirow{2}{*}{Self-} & \multirow{2}{*}{Synthetic}  & \multirow{2}{*}{Anomaly} & \multirow{2}{*}{Classif.} & \multicolumn{2}{c|}{Avenue} & \multicolumn{2}{c|}{Shanghai} \\
   \cline{6-9}
   \multirow{2}{*}{weights} & \multirow{2}{*}{distillation} & \multirow{2}{*}{data} & \multirow{2}{*}{maps} & \multirow{2}{*}{head} & \multicolumn{4}{c|}{AUC} \\
   \cline{6-9}
     &  &  &  &  & Micro & Macro & Micro & Macro\\
    \hline
    \hline
    & & & & & 84.0 & 85.6 & 69.7 & 80.1\\
    \hline
    \checkmark & & & & & 84.8 & 86.3 & 71.3 & 80.9\\
    \hline
    \checkmark& \checkmark & & & & 88.5 & 86.0 & 76.3 & 83.8\\
    \hline
    \checkmark& \checkmark & \checkmark & & & 88.5 & 86.9 & 77.0 & 83.0\\
    \hline
    \checkmark& \checkmark & \checkmark & \checkmark & & {90.5} & {89.6} & 77.3 & 84.2\\
    \hline
        \checkmark& \checkmark & \checkmark & \checkmark & \checkmark & \textbf{91.3} & \textbf{90.9} & \textbf{79.1} & \textbf{84.7}\\

    \hline
\end{tabular}
}
\vspace{-0.2cm}
\caption{Impact of each novel component on the micro and macro AUC scores (in \%), on Avenue \cite{Lu-ICCV-2013} and ShanghaiTech \cite{Luo-ICCV-2017}.}
\vspace{-0.1cm}
\label{tab_ablation_design} 
\end{table}

\begin{table}[t]
\centering 
\setlength\tabcolsep{1.5pt}
\mysize{
\begin{tabular}{| l | c | c | c | c |} 
\hline
  \multirow{3}{*}{Strategy} & \multicolumn{2}{c|}{Avenue} & \multicolumn{2}{c|}{Shanghai} \\
    \cline{2-5}
  &  \multicolumn{4}{c|}{AUC} \\
   \cline{2-5}
     & Micro & Macro & Micro & Macro\\
    \hline
    \hline
    Teacher & 84.0 & 85.6 & 74.8 & 82.3\\
    Teacher + Student & 85.4 & 85.8 & 75.1 & 82.1\\
    Teacher + Teacher-Student Difference & \textbf{88.5} & \textbf{86.0} & \textbf{76.3} & \textbf{83.8} \\
    Teacher + Student + Teacher-Student Difference & 86.9 & 85.8 & 75.8 & 83.6\\
    \hline
\end{tabular}
}
\vspace{-0.2cm}
\caption{Impact of strategies to combine the outputs of the teacher and student models on Avenue \cite{Lu-ICCV-2013} and ShanghaiTech \cite{Luo-ICCV-2017}. These results do not include the synthetic anomalies and the classification head.}
\vspace{-0.1cm}
\label{tab_ablation_distillation} 
\end{table}

\begin{table}[t]
\centering 
\mysize{
\begin{tabular}{| c | c | c | c | c | c |} 
\hline
   \multirow{2}{*}{Data set} & \multirow{2}{*}{Measure} & \multicolumn{4}{c|}{Percentage of synthetic data}\\
   \cline{3-6}
   & & $\;0\%\;$ & $\;25\%\;$ & $\;50\%\;$ & $\;75\%\;$\\
    \hline
    \hline
   \multirow{2}{*}{Avenue} & Micro AUC  & 88.5 & \textbf{91.3} & 90.6 & 89.9\\
   & Macro AUC  & 86.0 & \textbf{90.9} & 89.4 & 87.7\\
   \hline
   \multirow{2}{*}{Shanghai} & Micro AUC & 76.3 & \textbf{79.1} & 77.9 & 77.7 \\
   & Macro AUC  & 83.8 & \textbf{84.7} & 84.4 & 84.1 \\
    \hline
\end{tabular}
}
\vspace{-0.2cm}
\caption{Impact of varying the proportion of synthetic anomalies on the Avenue  \cite{Lu-ICCV-2013} and ShanghaiTech \cite{Luo-ICCV-2017} data sets.}
\vspace{-0.2cm}
\label{tab_proportion_anomalies} 
\end{table}

\begin{figure*}[!th]
\begin{center}
\centerline{\includegraphics[width=0.76\linewidth]{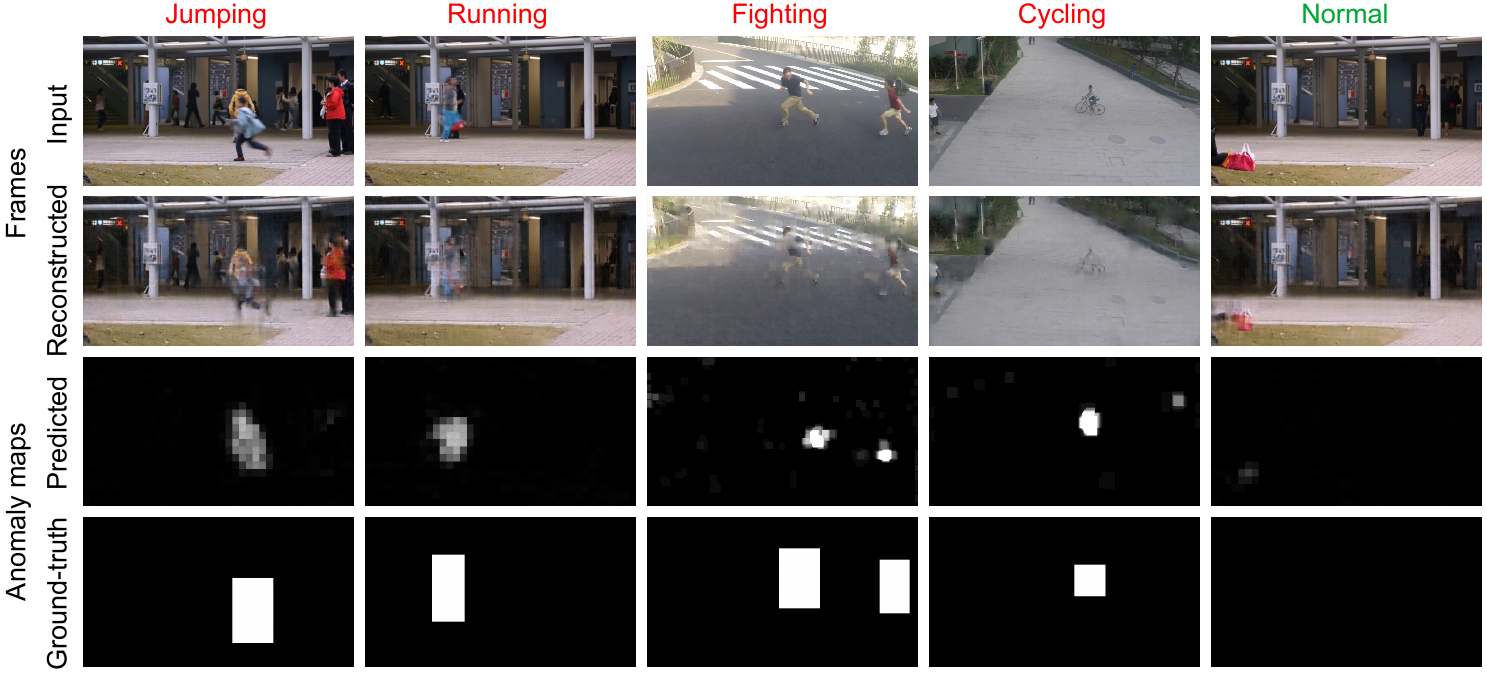}}
\vspace{-0.25cm}
\caption{Examples of frames and anomaly maps reconstructed by our teacher. The first four columns correspond to abnormal examples from the Avenue and ShanghaiTech data sets, while the last column shows a normal example. Best viewed in color.}
\label{fig_qualitative_results}
\vspace{-0.9cm}
\end{center}
\end{figure*}

\noindent
\textbf{Ablation study.}
In Table~\ref{tab_ablation_design}, we illustrate the impact of each novel component on our model's performance. The first model is a vanilla masked AE, which obtains a rather low performance on both Avenue and ShanghaiTech. Each and every component contributes towards boosting the performance of the vanilla model. We observe that self-distillation gives the highest boost in terms of the micro AUC. However, to surpass the $90\%$ milestone on Avenue, it is mandatory to introduce the prediction of anomaly maps in the learning task. An additional boost is given by our classification head.

The self-distillation procedure gives us a few possible strategies to combine the outputs of the teacher and the student. Hence, we investigate this aspect and report the results in Table~\ref{tab_ablation_distillation}. We observe that the best micro AUC is obtained when we combine the teacher reconstruction error with the teacher-student discrepancy. 

The proportion of synthetic examples per mini-batch is another aspect that can influence our model's performance. In Table~\ref{tab_proportion_anomalies}, we report the micro and macro AUC scores for three possible augmentation levels. We observe that augmentation is always useful, but, for a better outcome, it requires a moderate percentage ($25\%$).

\noindent
\textbf{Performance-speed trade-off.} In Figure~\ref{fig_tradeoff_av}, we compare our model with several other methods in terms of the performance-speed trade-off. This comparison undoubtedly shows that our method reaches a far better processing speed, while achieving fairly good performance. To strengthen this observation, we also compare the methods in terms of GFLOPs and number of parameters in Table~\ref{tab_speed_info}. We underline that the method of Gong \etal~\cite{Gong-ICCV-2019} might seem small in terms of the number of parameters, but it is slowed down by its input, which is formed of a cuboid constructed by stacking 16 consecutive frames. Moreover, the method relies on a memory module, where each memory slot records the features of one pixel in the activation maps. Although their method has twice as many parameters as our own, the large input volume and the sizable memory bank reduce the speed to 35 FPS.
With an FPS of 1655, our method proves to be significantly lighter than all its competitors.

\begin{table}[t]
\centering 
\setlength\tabcolsep{3.2pt}
\mysize{
\begin{tabular}{| l | c | c | c |} 
\hline
  Method & GFLOPs $\downarrow$ &  \#Params (M) $\downarrow$ & FPS $\uparrow$\\
\hline
\hline
    Georgescu \etal~\cite{Georgescu-CVPR-2021} & 107.9 & 65 & 51 \\
    Georgescu \etal~\cite{Georgescu-TPAMI-2021} & 121.6 & 67 & 24\\
    Liu \etal~\cite{Liu-ICCV-2021} & 179.5 & 320 & 12\\
    Park \etal~\cite{Park-WACV-2022} & 84 & 64 & 195\\
    Gong \etal~\cite{Gong-ICCV-2019} & 55.2 & 6 & 35\\
\hline
    Ours & \textbf{0.8} & \textbf{3} & \textbf{1655} \\
\hline
\end{tabular}
}
\vspace{-0.2cm}
\caption{Comparing methods in terms of floating point operations (GFLOPs), number of parameters, and FPS.}
\label{tab_speed_info} 
\vspace{-0.2cm}
\end{table}

\noindent
\textbf{Qualitative results.} In Figure~\ref{fig_qualitative_results}, we present the frames and anomaly maps reconstructed by the teacher in four abnormal scenarios from Avenue and ShanghaiTech. Moreover, in the fifth column, we illustrate the behavior of the teacher in a normal scenario. In all four abnormal cases, the reconstruction error is visibly higher for anomalous regions. This effect is mostly due to our training procedure based on synthetic data augmentation. The most obvious example is in the fourth column, where the bicycle seen in a pedestrian area is almost entirely removed from the output. Additionally, the predicted anomaly maps are well aligned with the ground-truth ones. 

\vspace{-0.15cm}
\section{Conclusion}
\vspace{-0.1cm}

In this work, we proposed a lightweight masked auto-encoder (3M parameters, 0.8 GFLOPs) for video anomaly detection, which learns to reconstruct tokens with high motion gradients. Our framework is based on self-distillation, leveraging the discrepancy between teacher and student decoders for anomaly detection. Moreover, we boost the performance of our model by introducing a data augmentation technique based on overlapping synthetic anomalies on normal training data. Our highly efficient framework reached an unprecedented speed of 1655 FPS, with a minimal performance gap with respect to the state-of-the-art object-centric approaches. 


\noindent
\textbf{Acknowledgments.}
Work supported by a grant of the Romanian Ministry of Education and Research, CNCS-UEFISCDI, project number PN-III-P2-2.1-PED-2021-0195, within PNCDI III.

{\small
\bibliographystyle{ieeenat_fullname}
\bibliography{references}
}

\section{Supplementary}

In the supplementary, we present localization results, as well as additional ablation and qualitative results. Finally, we discuss the connections between our approach and other frameworks based on masked auto-encoders.

\subsection{Additional Results}

\begin{figure}[!t]
\begin{center}
\centerline{\includegraphics[width=1.0\linewidth]{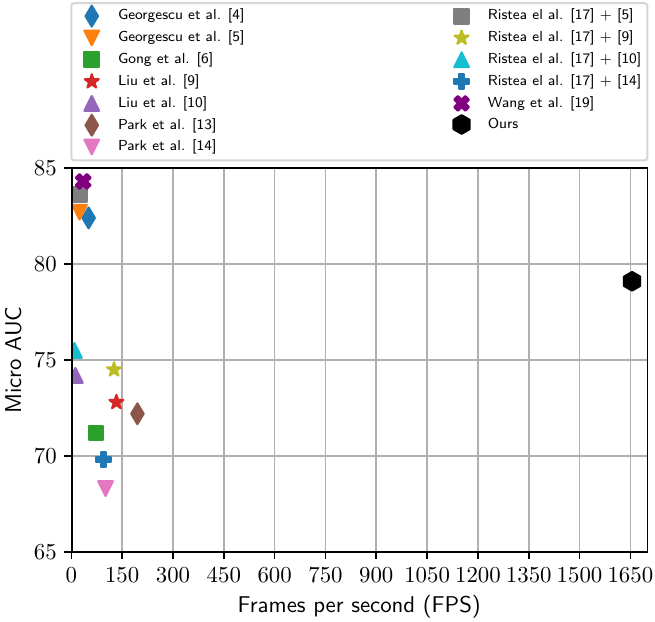}}
\vspace{-0.25cm}
\caption{Performance versus speed trade-offs for our self-distilled masked AE and several state-of-the-art methods \cite{Georgescu-CVPR-2021,Georgescu-TPAMI-2021,Ristea-CVPR-2022,Liu-CVPR-2018,Liu-ICCV-2021,Park-CVPR-2020,Gong-ICCV-2019,Wang-ECCV-2022,Park-WACV-2022} (with open-sourced code), on the ShanghaiTech data set. The running times of all methods are measured on a computer with one Nvidia GeForce GTX 3090 GPU with 24 GB of VRAM. Best viewed in color.}
\label{fig_tradeoff_sh}
\vspace{-0.9cm}
\end{center}
\end{figure}

\begin{figure*}[!th]
\begin{center}
\centerline{\includegraphics[width=1\linewidth]{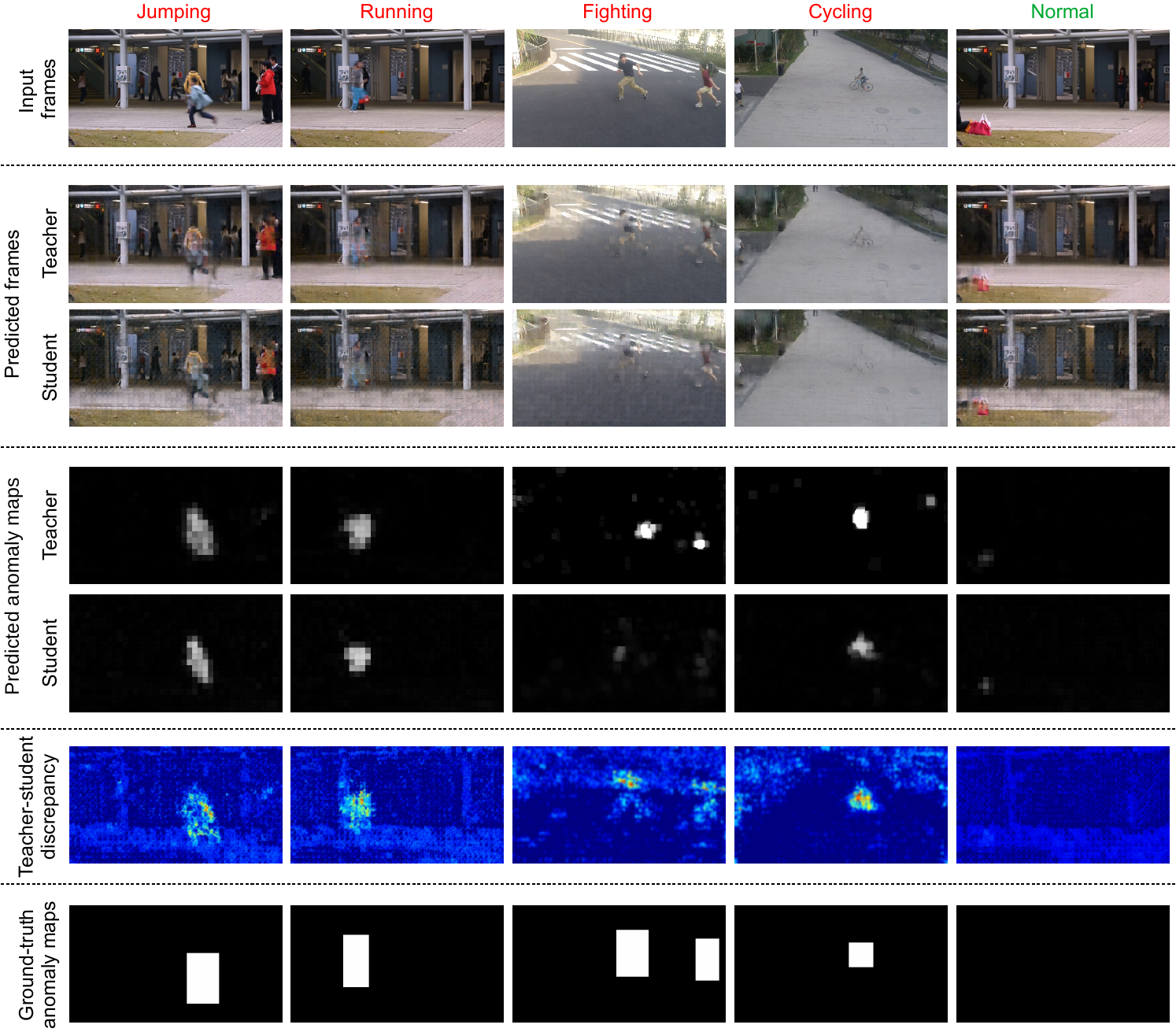}}
\vspace{-0.15cm}
\caption{Examples of frames and anomaly maps reconstructed by our teacher and student models. Additionally, the differences (discrepancy maps) between the teacher and student outputs are shown in the sixth row. The first four columns correspond to abnormal examples from the Avenue and ShanghaiTech data sets, while the last column corresponds to a normal example. Best viewed in color.}
\label{fig_qualitative_results}
\vspace{-0.6cm}
\end{center}
\end{figure*}

\begin{figure}[!t]
\begin{center}
\centerline{\includegraphics[width=1.0\linewidth]{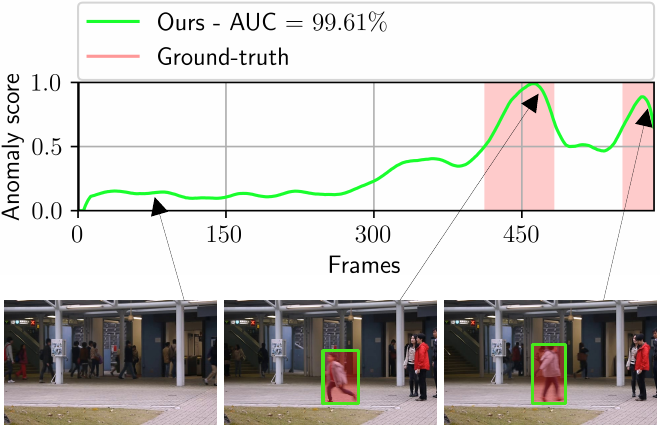}}
\vspace{-0.25cm}
\caption{Predictions for test video $07$ from Avenue. The abnormal bounding boxes are given by the convex hull of the patches labeled as abnormal. Best viewed in color.}
\label{fig_avenue_example2}
\vspace{-0.5cm}
\end{center}
\end{figure}

\begin{figure}[!t]
\begin{center}
\centerline{\includegraphics[width=1.0\linewidth]{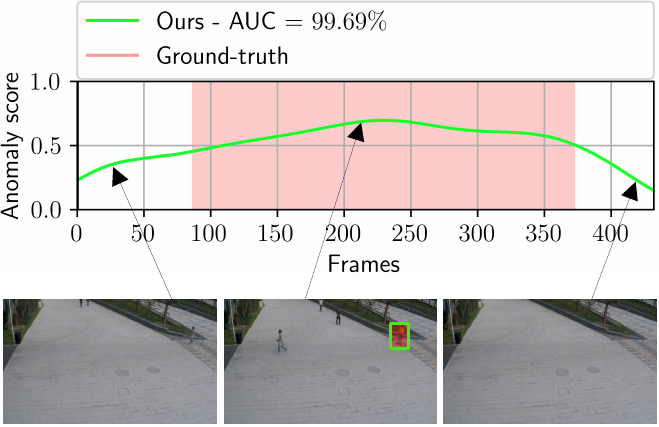}}
\vspace{-0.25cm}
\caption{Predictions for test video $01\_0015$ from ShanghaiTech. The abnormal bounding boxes are given by the convex hull of the patches labeled as abnormal. Best viewed in color.}
\label{fig_sh_example2}
\vspace{-0.9cm}
\end{center}
\end{figure}

\begin{figure}[!t]
\begin{center}
\centerline{\includegraphics[width=1.0\linewidth]{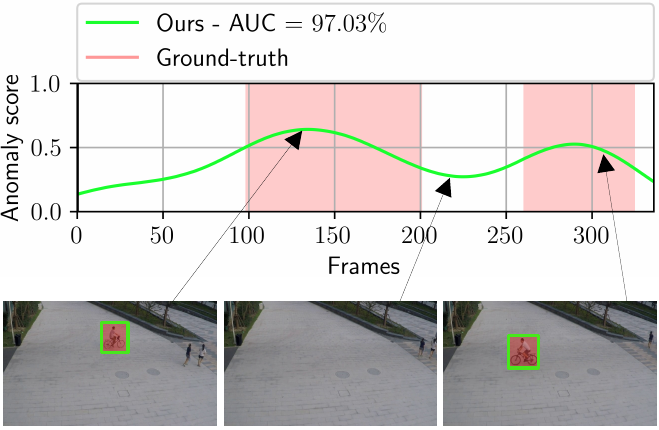}}
\vspace{-0.25cm}
\caption{Predictions for test video $01\_0051$ from ShanghaiTech. The abnormal bounding boxes are given by the convex hull of the patches labeled as abnormal. Best viewed in color.}
\label{fig_sh_example}
\vspace{-0.9cm}
\end{center}
\end{figure}

\begin{figure}[!t]
\begin{center}
\centerline{\includegraphics[width=1.0\linewidth]{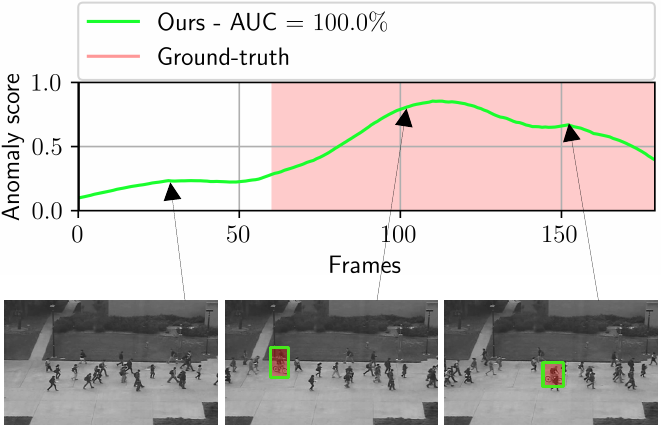}}
\vspace{-0.25cm}
\caption{Predictions for video Test001 from UCSD Ped2. The abnormal bounding boxes are given by the convex hull of the patches labeled as abnormal. Best viewed in color.}
\label{fig_ped2_example}
\vspace{-0.8cm}
\end{center}
\end{figure}

\noindent
\textbf{Performance-speed trade-off.} In the main article, we compared the performance-speed trade-off of our masked AE with other state-of-the-art methods on the Avenue data set. To demonstrate that our superior trade-off is maintained across data sets, we hereby analyze the trade-offs of several methods, including our own, on the ShanghaiTech data sets. The results illustrated in Figure \ref{fig_tradeoff_sh} clearly indicate that our method is significantly faster than competing methods, while surpassing the other frame-level anomaly detection methods. This observation confirms the consistency of our trade-off across data sets.

\begin{table}[t]
\centering 
\setlength\tabcolsep{2.2pt}
\small{
\begin{tabular}{| c | c | c | c | c | c | c | c | c |} 
\hline
 \multirow{3}{*}{\rotatebox[origin=c]{90}{\hspace{-0.3cm}Type}} &
 \multirow{3}{*}{\vspace{-0.3cm}Method} & \multicolumn{2}{c|}{Avenue} & \multicolumn{2}{c|}{Shanghai} & \multicolumn{2}{c|}{UBnormal}  &
 \multirow{3}{*}{\vspace{-0.3cm}FPS} \\
 \cline{3-8}
 & & \rotatebox[origin=c]{90}{$\;$RBDC$\;$} & \rotatebox[origin=c]{90}{TBDC} & \rotatebox[origin=c]{90}{RBDC} & \rotatebox[origin=c]{90}{TBDC} & \rotatebox[origin=c]{90}{RBDC} & \rotatebox[origin=c]{90}{TBDC} & \\
 \hline 
 \hline
 
 {\multirow{10}{*}{\rotatebox[origin=c]{90}{Object-centric}}} 
&
\cite{Barbalau-CVIU-2023} & 47.83 & 85.26 & {\color{ForestGreen}47.14} & {\color{ForestGreen}85.61} & {\color{red}25.63} & {\color{red}63.53} & 20\\
& 
\cite{Georgescu-CVPR-2021} &  57.00 & 58.30 & 42.80 & 83.90 & {\color{RoyalBlue}19.71} & {\color{RoyalBlue}55.80} & 51 \\
& 
\cite{Georgescu-TPAMI-2021} & {\color{RoyalBlue}65.05} & 66.85 & 41.34 & 78.79 & {\color{ForestGreen}25.43} & {\color{ForestGreen}56.27} & 24 \\
&
 \cite{Ionescu-CVPR-2019} & 15.77 & 27.01 & 20.65 & 44.54 & - & - & -\\
 &
\cite{Liu-ICCV-2021} & 41.05 & {\color{RoyalBlue}86.18} & 44.41 & 83.86 & - & - & 12 \\ 
&
\cite{Madan-ARXIV-2022} + \cite{Barbalau-CVIU-2023} & 49.01 & 85.94 &  {\color{red}47.73} & {\color{red}85.68} & - & - & 20\\
&
\cite{Madan-ARXIV-2022} + \cite{Georgescu-TPAMI-2021} & {\color{red}66.04} & 65.12 & 40.52 & 81.93 & - & - & 31\\
&
\cite{Madan-ARXIV-2022} + \cite{Liu-ICCV-2021} & 46.49 & {\color{ForestGreen}86.43} & 45.86 &  {\color{RoyalBlue}84.69} & - & - & 10 \\ 
&
\cite{Ristea-CVPR-2022} + \cite{Georgescu-TPAMI-2021} & {\color{ForestGreen}65.99} & 64.91 & 40.55 & 83.46 & - & - & 31 \\
&
\cite{Ristea-CVPR-2022} + \cite{Liu-ICCV-2021} & 62.27 & {\color{red}89.28} &  {\color{RoyalBlue}45.45} & 84.50 & - & - & 10\\ 

\hline
{\multirow{8}{*}{\rotatebox[origin=c]{90}{Frame or cube level}}}

& \cite{Bertasius-ICML-2021} & - & - & - & - & {\color{ForestGreen}0.04} & {\color{ForestGreen}0.05} & 37 \\
&
\cite{Liu-CVPR-2018} & 19.59 & 56.01 & 17.03 & 54.23 & - & - & 28\\
&
\cite{Madan-ARXIV-2022} + \cite{Liu-CVPR-2018} & 23.79 & 66.03 & {\color{ForestGreen}19.13} & {\color{ForestGreen}61.65} & - & - & 26\\
&
\cite{Ramachandra-WACV-2020a} & {\color{RoyalBlue}35.80} & {\color{red}80.90} & - & - & - & - & -\\
&
\cite{Ramachandra-WACV-2020b} & {\color{ForestGreen}41.20} & {\color{ForestGreen}78.60} & - & - & - & - & -\\
&
\cite{Ristea-CVPR-2022} + \cite{Liu-CVPR-2018} & 20.13 & 62.30 & {\color{RoyalBlue}18.51} & {\color{RoyalBlue}60.22} & - & - & 26\\
&
\cite{Sultani-CVPR-2018} & - & - & - & - & {\color{RoyalBlue}0.01} & {\color{RoyalBlue}0.01} & 56 \\
\cline{2-9}
& Ours & {\color{red}46.77} & {\color{RoyalBlue}66.58}  & {\color{red}26.42} & {\color{red}66.67} & {\color{red}23.58} & {\color{red}50.36} & 1655\\
\hline
\end{tabular}
}
\vspace{-0.2cm}
\caption{RBDC and TBDC scores (in \%) of several state-of-the-art frame-level, cube-level and object-level methods versus our self-distilled masked AE on Avenue, ShanghaiTech and UBnormal. The top three scores for each category of methods are shown in {\color{red}red}, {\color{ForestGreen}green}, and {\color{RoyalBlue}blue}. All reported running times (including those of the baselines) are measured on a machine with an Nvidia GeForce GTX 3090 GPU with 24 GB of VRAM.}
\vspace{-0.3cm}
\label{tab_results_rbdc_tbdc} 
\end{table}

\noindent
\textbf{Anomaly localization results.} To measure anomaly localization performance, we employ the recently proposed Region-Based Detection Criterion (RBDC) and Track-Based Detection Criterion (TBDC) \cite{Ramachandra-WACV-2020a}. Following Ramachandra \etal \cite{Ramachandra-WACV-2020a}, we set the region overlap threshold to $0.1$ and the track overlap threshold to $0.1$, which allows us to directly compare with other methods reporting the RBDC and TBDC scores. In Table \ref{tab_results_rbdc_tbdc}, we report the RBDC and TBDC scores of our method versus frame-level and object-centric methods, on the Avenue, ShanghaiTech and UBnormal data sets.

When compared with frame-level and cube-level methods, our approach obtains the best RBDC scores on all three data sets. Furthermore, our method outperforms all other frame-level and cube-level methods on ShanghaiTech and UBnormal, in terms of TBDC. The most dramatic differences in favor of our method are reported on the UBnormal data set. Notably, our method also outperforms some of the object-centric approaches, in terms of both RBDC and TBDC. Considering that our approach is a frame-level method, its anomaly localization results are remarkable. Not only that our method is generally better than frame-level and cube-level methods in terms of both RBDC and TBDC, but its processing speed is significantly higher.

\begin{figure}[th!]
\centering
\begin{subfigure}{0.98\linewidth}
    \includegraphics[width=1.0\linewidth]{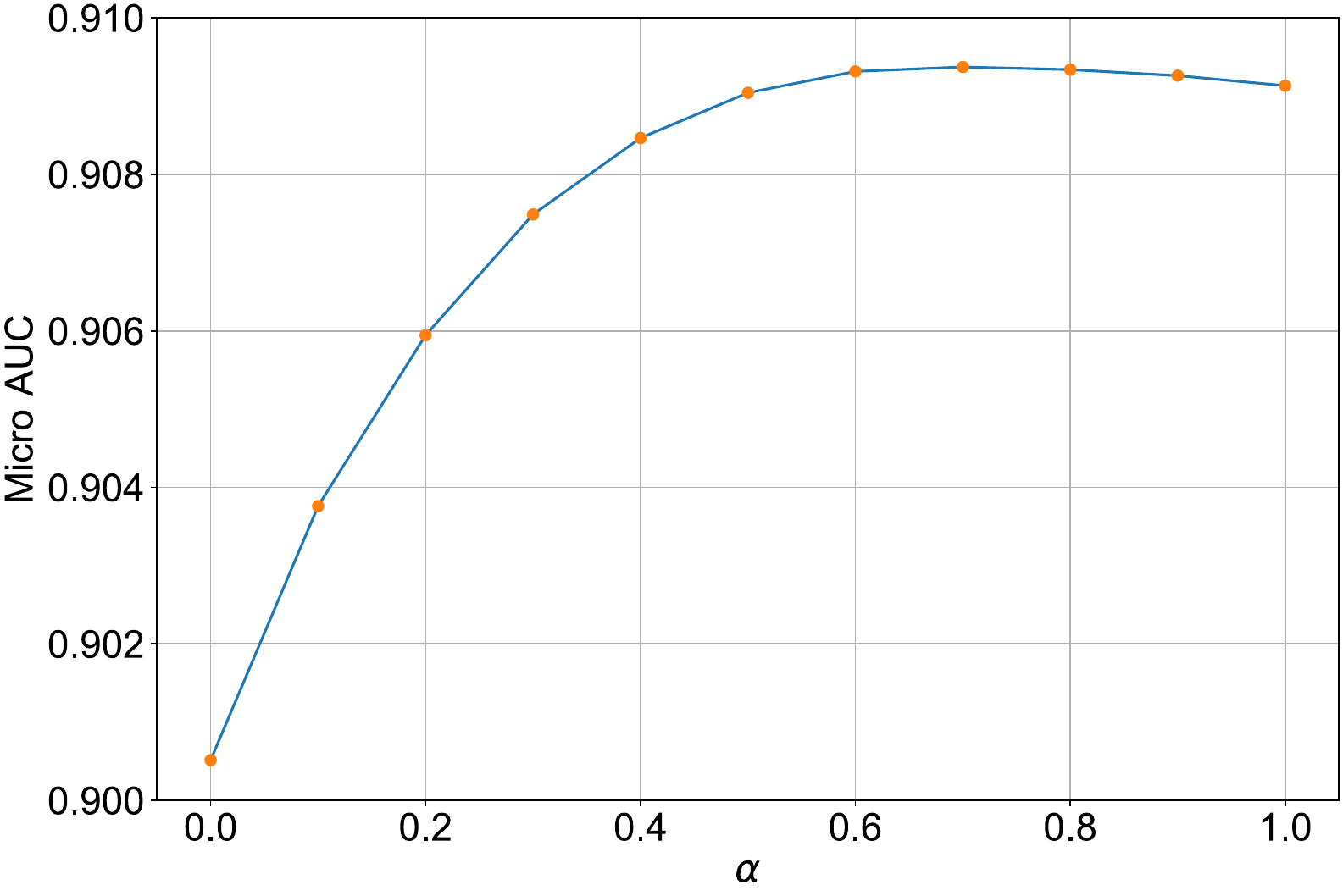}
    \vspace{-0.5cm}
    \caption{Varying $\alpha$, while keeping $\beta=0.5$ and $\gamma=0.5$.}
    \label{fig_abg:first}
     \vspace{0.2cm}
\end{subfigure}
\begin{subfigure}{0.98\linewidth}
    \includegraphics[width=1.0\linewidth]{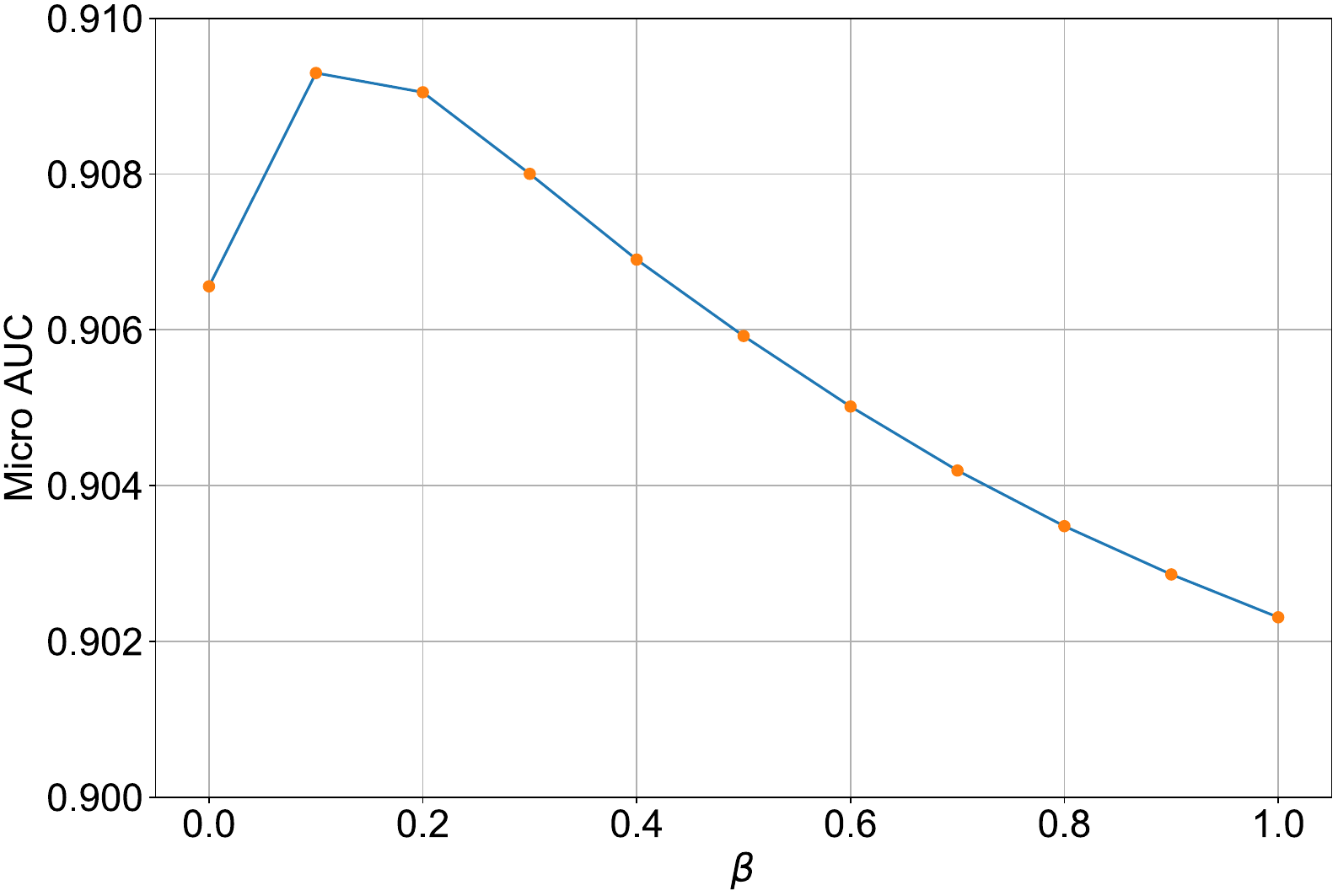}
    \vspace{-0.5cm}
    \caption{Varying $\beta$, while keeping $\alpha=0.5$ and $\gamma=0.5$.}
    \label{fig_abg:second}
        \vspace{0.2cm}
\end{subfigure}
\begin{subfigure}{0.98\linewidth}
    \includegraphics[width=1.0\linewidth]{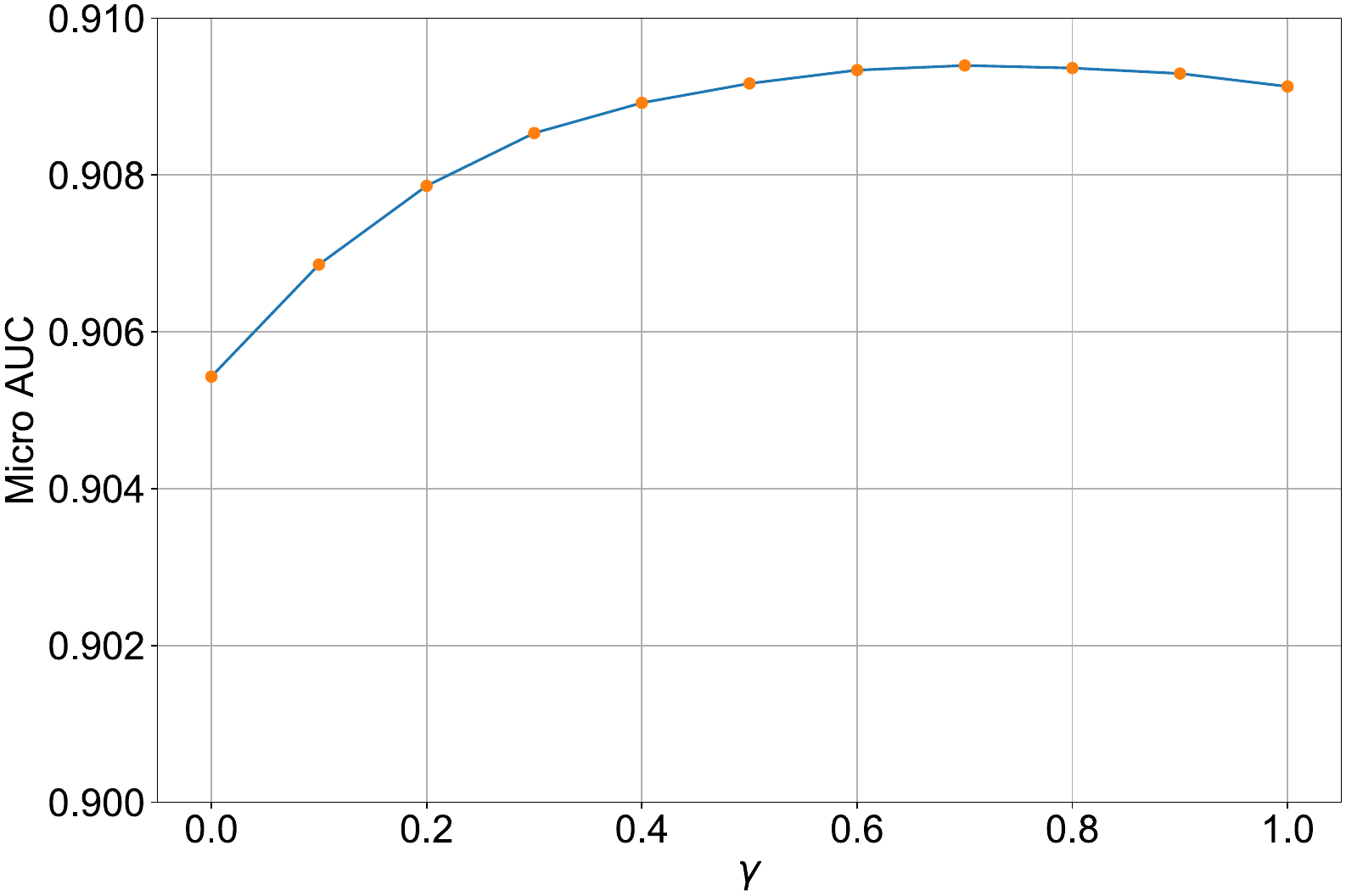}
    \vspace{-0.5cm}
    \caption{Varying $\gamma$, while keeping $\alpha=0.5$ and $\beta=0.5$.}
    \label{fig_abg:third}
\end{subfigure}
\vspace{-0.1cm}
\caption{Micro AUC scores on the Avenue data set, while varying the hyperparameters $\alpha$, $\beta$ and $\gamma$ controlling the anomaly score contributions of the teacher decoder, the teacher-student discrepancy, and the classification head, respectively. Each hyperparameter is varied between $0$ and $1$, while keeping the others fixed to $0.5$.}
\label{fig_alpha_beta_gamma}
\vspace{-0.3cm}
\end{figure}

\noindent
\textbf{Qualitative results.}
In Figure~\ref{fig_qualitative_results}, we illustrate the frame reconstructions and the anomaly maps returned by the teacher and student models for five input frames. We keep the same five examples as in the main paper, essentially adding the outputs from the student model, as well as the discrepancy maps between the teacher and the student. For the first four examples, which are abnormal, we can see that the frame reconstructions of both teacher and student models are deficient in the anomalous regions, as desired. Moreover, in the fourth example, the student entirely removes the bicycle from its reconstructed output, which triggers a true positive detection. The anomaly maps generated by the teacher are generally better than the ones generated by the student. The latter maps are well aligned with the ground-truth anomalies, but the predicted anomalies cover a smaller than expected area. However, the discrepancy maps exhibit intense disagreements in the anomalous regions, indicating that the discrepancy maps are good indicators for abnormal events. For the normal example depicted in the fifth column, the anomaly and discrepancy maps do not show any pixels with high anomaly scores, confirming that our method yields the desired effect. 

Another interesting remark is that the reconstructed frames returned by the student are worse than those of the teacher. This happens because the student learns to reconstruct the teacher's output frames instead of the original input frames. Nevertheless, the reconstruction power of the student is less important to us, \ie we care more about obtaining discrepancy maps that are highly correlated with the abnormal events. As discussed above, our student works as expected, helping the teacher to better predict the anomalies.

In Figure \ref{fig_avenue_example2}, we illustrate the anomaly scores for test video $07$ from the Avenue data set. On this test video, our model reaches an AUC higher than $99\%$, being able to accurately identify the person running and jumping around.

In Figure \ref{fig_sh_example2}, we showcase the anomaly scores for video $01\_0015$ from the ShanghaiTech test set. As in the previous example, our model obtains an AUC higher than $99\%$, returning higher anomaly scores when the skateboarder passes through the pedestrian area.

In Figure \ref{fig_sh_example}, we present the anomaly scores for video $01\_0051$ from the ShanghaiTech test set. Our model reaches an AUC of $97.03\%$ on this video, being able to flag and locate the abnormal event, namely riding a bike into a pedestrian area.

In Figure \ref{fig_ped2_example}, we illustrate the anomaly scores for video Test001 from UCSD Ped2. Here, our model reaches an AUC of $100\%$, being able to perfectly differentiate between normal and abnormal events.

\noindent
\textbf{Ablating pointwise convolutions.} We next assess the impact of replacing the fully connected layers inside the vanilla CvT blocks \cite{Wu-ICCV-2021} with pointwise convolutions. The results presented in Table \ref{tab_ablation_pointwise} show that our minor architectural change leads to a speed boost of 211 FPS and an increase of $2.1\%$ in terms of the micro AUC. The results confirm that the pointwise convolutions provide a superior trade-off between accuracy and speed.

\noindent
\textbf{Ablating anomaly score components.} In Figure~\ref{fig_alpha_beta_gamma}, we illustrate the impact of $\alpha$, $\beta$, and $\gamma$ on the micro AUC score computed on the Avenue data set. These hyperparameters are the weights associated to the three anomaly score components, namely the teacher decoder, the teacher-student discrepancy, and the classification head. We note that all weight configurations lead to micro AUC scores higher than $90\%$, indicating that our method is fairly robust to suboptimal tuning of $\alpha$, $\beta$, and $\gamma$. Indeed, the vast majority of combinations lead to micro AUC scores that are higher than the micro AUC scores of all other frame-level and cube-level methods evaluated on Avenue (see Table \ref{tab_results}). Nonetheless, we generally observe that the teacher decoder and the classification head should have higher weights than the teacher-student discrepancy.

\begin{table}[t]
\centering 
\setlength\tabcolsep{2.7pt}
\small{
\begin{tabular}{| l | c | c | c |} 
\hline
  \multirow{2}{*}{CvT block type} &  \multicolumn{2}{c|}{AUC} & \multirow{2}{*}{FPS} \\
   \cline{2-3}
     & Micro & Macro & \\
    \hline
    \hline
    MLP \cite{Wu-ICCV-2021} & $89.2$ & $88.1$ & 1454 \\
    Pointwise convolutions (ours) & $91.3$ & $90.9$ & 1655\\
    \hline
\end{tabular}
}
\vspace{-0.2cm}
\caption{Micro and macro AUC scores (in \%) on Avenue \cite{Lu-ICCV-2013} with pointwise convolutional layers versus fully connected layers in the CvT transformer blocks.}
\vspace{-0.1cm}
\label{tab_ablation_pointwise} 
\end{table}

\subsection{Extended Related Work}

Driven by the goal of learning better high-level representations, some studies, such as \cite{Chen-ECCV-2022,Yang-ARXIV-2022}, tried to modify the pretraining phase of the masked AE \cite{He-CVPR-2022}. Since these methods \cite{Chen-ECCV-2022,Yang-ARXIV-2022} may appear to be related to our approach, we discuss the differences in detail below.

Chen \etal~\cite{Chen-ECCV-2022} argued that the pretraining procedure of the vanilla masked AE \cite{He-CVPR-2022} is suboptimal because learning to reconstruct low-level information is not necessarily beneficial for tasks such as classification. Hence, they propose a procedure to reconstruct the high-level representations of the masked tokens instead. The training is performed by maximizing the cosine similarity between teacher and student representations. The teacher is an encoder given by the exponential moving average of past versions of the student encoder. In their case, this training process is called self-distillation because the student learns from aggregated past versions of itself. In our case, self-distillation refers to the fact that the teacher and the student have a shared (identical) encoder. Hence, there is a large difference in terms of the architecture and the training procedure between our model and that of Chen \etal~\cite{Chen-ECCV-2022}. This is also confirmed by the fact that Chen \etal~\cite{Chen-ECCV-2022} does not even cite the work of Zhang \etal~\cite{Zhang-PAMI-2022}, which introduces the form of self-distillation that inspired our work.

Yang \etal~\cite{Yang-ARXIV-2022} modified the vanilla  masked AE to learn a spatio-temporal representation. The architecture attaches an additional decoder, which is trained to reconstruct the motion gradients. Unlike Yang \etal~\cite{Yang-ARXIV-2022}, we do not attempt to reconstruct the motion gradients. Instead, we leverage the motion gradient information to make our model focus on reconstructing tokens which correspond to higher motion. This is necessary to avoid reconstructing the static background scene, which is predominant in anomaly detection data sets.

Aside from the technical differences, another aspect that creates an even higher gap between our method and those of Chen \etal~\cite{Chen-ECCV-2022} and Yang \etal~\cite{Yang-ARXIV-2022} is the target task. Indeed, our masked AE is specifically designed for abnormal event detection in video, while the masked AEs proposed in \cite{Chen-ECCV-2022,Yang-ARXIV-2022} are focused on improving the pretraining procedure.

\end{document}